\documentclass{article}

\PassOptionsToPackage{numbers, compress}{natbib}
\usepackage[preprint]{neurips_2020}
\usepackage[utf8]{inputenc} 
\usepackage[T1]{fontenc}    
\usepackage{hyperref}   
\usepackage{url}            
\usepackage{booktabs}       
\usepackage{amsfonts}       
\usepackage{nicefrac}       
\usepackage{microtype}      
\usepackage{graphicx}
\usepackage{amsmath}
\usepackage{subcaption}
\usepackage[ruled]{algorithm2e}
\usepackage{textcomp}
\makeatletter
\g@addto@macro{\UrlBreaks}{\UrlOrds}

\usepackage{xcolor}
\usepackage{color}

\title{On Lyapunov Exponents for RNNs: \\Understanding Information Propagation Using Dynamical Systems Tools}

\author{
  Ryan Vogt$^*$\\
  Department of Applied Mathematics\\
  University of Washington\\
  \texttt{ravogt95@uw.edu}
      \And
   Maximilian Puelma Touzel$^*$ \\
      Math \& Stats Dept., Universit\'e de Montr\'eal\\
  Mila, Qu\'ebec AI Institute \\
  \texttt{puelmatm@mila.quebec}
   \And
   Eli Shlizerman$^{**}$ \\
   Depts. of App. Math. and Elec. $\&$ \\ Comp. Engineering,
   University of Washington\\ 
     \texttt{shlizee@uw.edu}
     \And
   Guillaume Lajoie$^{**}$ \\
    Math \& Stats Dept., Universit\'e de Montr\'eal\\
     Mila, Qu\'ebec AI Institute  \\
   \texttt{g.lajoie@umontreal.ca}
}

\begin{document}

\maketitle

\begin{center}
$*$: authors share first authorship, order selected at random. \\
$**$: authors share senior authorship, order selected at random. 
\end{center}

\begin{abstract}
Recurrent neural networks (RNNs) have been successfully applied to a variety of problems involving sequential data, but their optimization is sensitive to parameter initialization, architecture, and optimizer hyperparameters.
Considering RNNs as dynamical systems, a natural way to capture stability, i.e., the growth and decay over long iterates, are the Lyapunov Exponents (LEs), which form the Lyapunov spectrum. 
The LEs have a bearing on stability of RNN training dynamics because forward propagation of information is related to the backward propagation of error gradients. 
LEs measure the asymptotic rates of expansion and contraction of nonlinear system trajectories, and generalize stability analysis to the time-varying attractors structuring the non-autonomous dynamics of data-driven RNNs.
As a tool to understand and exploit stability of training dynamics, the Lyapunov spectrum fills an existing gap between prescriptive mathematical approaches of limited scope and computationally-expensive empirical approaches.
To leverage this tool, we implement an efficient way to compute LEs for RNNs during training, discuss the aspects specific to standard RNN architectures driven by typical sequential datasets, and show that the Lyapunov spectrum can serve as a robust readout of training stability across hyperparameters.
With this exposition-oriented contribution, we hope to draw attention to this under-studied, but theoretically grounded tool for understanding training stability in RNNs.
\end{abstract}

\vspace*{-1mm}
\section{Introduction}
\vspace*{-1mm}
The propagation of error gradients in deep learning leads to the study of recursive compositions and their stability~\cite{Bengio1994}.
Vanishing and exploding gradients arise from long products of Jacobians of the hidden state dynamics whose norm exponentially grows or decays, hindering training~\cite{Pascanu}. 
To mitigate this sensitivity, much effort has been made to mathematically understand the link between model parameters and the the eigen- and singular-value spectra of these long products~\cite{Chen2018,Pennington2017,Poole2016}.
For architectures used in practice, this approach appears to have limited a scope~\cite{Yang2017,zheng2020r} likely due to spectra having non-trivial properties reflecting complicated long-time dependencies within the trajectory. 
Fortunately, the theory of dynamical systems has formulated this problem for general dynamical systems, i.e., for arbitrary architectures. 
The Lyapunov spectrum (LS) formed by the Lyapunov exponents (LEs) is precisely the measurement that characterizes rates of expansion and contraction of nonlinear system dynamics.

While there are results using LEs in machine learning~\cite{Legenstein:2007th,Pennington:2018tg,Laurent:2016vs}, most only look at the largest exponent, the sign of which indicates the presence or absence of chaos. Additionally, to our knowledge, current machine learning uses of LEs are for the autonomous case,  i.e. , analyzing RNN dynamics in the absence of inputs, a regime at odds with the most common use of RNNs as input-processing systems.
Thus, there is likely much to be gained from both the study of other features of the LS and from including input-driven regimes into LEs analysis. Indeed, a concurrent work coming from theoretical neuroscience takes up this approach \cite{Engelken} and outlines the ongoing work using LEs to study models of neural networks in the brain. However, the machinery to compute the LS and the theory surrounding its interpretation is still relatively unknown in the machine learning community. 

In this exposition-style paper, we aim to fill this gap by presenting an overview of LEs in the context of RNNs, and discuss their usefulness for studying training dynamics. We present encouraging results which suggest that LEs could serve in efficient hyperparameter tuning by allowing early detection of performance. 
To compute LEs, we present a novel algorithm for various RNN architectures, taking advantage of modern deep learning environments. 
Finally, we highlight key directions of research that have the potential to leverage LEs for improvement of RNNs robustness and performance.

\vspace*{-1mm}
\section{Motivation and Definitions}
\vspace*{-1mm}
Here we develop the problem of spectral constraints for robust gradient propagation. For transparent exposition, in this section we will consider the "vanilla" RNN defined as
\begin{align}
    \mathbf{o}_t=\mathbf{W}\mathbf{h}_t,\;\mathbf{h}_{t}&=\phi(\mathbf{a}_t),\; \mathbf{a}_t=\mathbf{V}\mathbf{h}_{t-1}+\mathbf{U}\mathbf{x}_{t}+\mathbf{b}\;,\label{eq:vanilla}
\end{align}
where $\mathbf{V}$ is the recurrent weight matrix that couples elements of the hidden state vector $\mathbf{h}_t\in\mathbb{R}^{N}$, $\mathbf{U}$ projects the input, $\mathbf{x}_t$, into the network, $\mathbf{b}$ is a constant bias vector, and $\phi$ applies a nonlinear scalar transformation element-wise. The readout weights, $\mathbf{W}$, output the activity into the output variable, $\mathbf{o}_t$. The loss over $T$ iterates is $L=\sum_{t=1}^{T}L_t$, with $L_t=f(\mathbf{y}_t,\hat{\mathbf{y}}_t)$, with $f$ some scalar loss function (e.g., cross-entropy loss,  $f(\mathbf{y},\hat{\mathbf{y}})={\mathbf{y}}\cdot\log\hat{\mathbf{y}}$), $\hat{\mathbf{y}}_t$ the prediction (e.g., $\hat{\mathbf{y}_t}=\mathrm{softmax}(\mathbf{o}_t)$), and $\mathbf{y}_t$ is a one-hot binary target vector. The parameters, $\Theta=(\mathbf{V},\mathbf{U},\mathbf{b},\mathbf{W})$, are learned by following the gradient of the loss, {\it e.g.} in the space of recurrent weights $\mathbf{V}$,
\begin{align}
    \nabla_{\mathbf{V}}L=\sum_{t=1}^T\sum_{i=1}^N \frac{\partial L}{\partial h_{t,i}}\nabla_\mathbf{V}h_{t,i}=\sum_{t=1}^T \textrm{diag}(\phi'(\mathbf{a}_t))\nabla_{\mathbf{h}_t}L \;\mathbf{h}_{t-1}^\top\;,
\end{align}
where $\textrm{diag}(\mathbf{x})$ is the diagonal matrix formed by the vector $\mathbf{x}$ and $\phi'$ is the derivative of $\phi$ and $^\top$ denotes transpose. Here,
\begin{align}
    \nabla_{\mathbf{h}_t}L=\sum_{s=t}^T\left(\prod_{r=t+1}^s \mathbf{J}^\top_r\right)\mathbf{W}^\top\nabla_{\mathbf{o}_s}L\;,\label{eq:gradloss}
\end{align}
where $\nabla_{\mathbf{o}_s}L$ is some simple expression ({\it e.g.} $\hat{\mathbf{y}}-\mathbf{y}_t$ for cross-entropy loss) and  $\mathbf{J}_t=\frac{\partial \mathbf{h}_{t}}{\partial \mathbf{h}_{t-1}}$ is the Jacobian of the hidden state dynamics,
\begin{align}
    \mathbf{J}_{t}&=\textrm{diag}\left(\phi'(\mathbf{a}_t)\right)\mathbf{V}\;.
\end{align}
$\mathbf{J}_t$ varies in time with $\mathbf{x}_t$ and $\mathbf{h}_{t-1}$ via $\mathbf{a}_t$ and so is treated as a random matrix with ensemble properties arising from the specified input statistics and the emergent hidden state statistics. 

Gradients can vanish or explode with $T$ according to the singular value spectra of the products of Jacobians appearing in Eq.~\eqref{eq:gradloss}~\cite{Bengio1994}. This fact motivated the study of hyperparameter constraints that control the average of the latter over the input distribution. Specifically, the scale parameter of Gaussian or orthogonal initializations of $\mathbf{V}$ is chosen such that the condition number (ratio of largest to smallest singular value) of $\mathbf{J}_t$ is bounded around 1 for all $t$ \cite{Pascanu}, or such that the first and second moments of the distribution of squared singular values (averaged over inputs) of the Jacobian products are chosen near 1 and 0, respectively according to dynamical isometry~\cite{Pennington2017}. Parameter conditions for the latter have been derived for i.d.d. Gaussian input. The approach has been extended to RNNs under the assumption of untied weights with good empirical correspondence in vanilla and minimalRNN~\cite{Chen2018}, but larger discrepancy in LSTMs~\cite{Gilboa}, suggesting there are other contributions to stability in more complex, state-of-the-art architectures~\cite{Yang2017,zheng2020r}. 

Better understanding how robust gradient propagation emerges in these less idealized settings that include high-dimensional and temporally correlated input sequences demands a more general theory. These Jacobian products appearing in Eq.~\eqref{eq:gradloss} can each be expressed as the transpose of a forward sequence, $\prod_{r=t+1}^s \mathbf{J}^\top_r=\left( \mathbf{J}_{s}\cdots\mathbf{J}_{t+1}\right)^\top$. Note that the backward time gradient dynamics shares properties with the forward time hidden state dynamics. This is an important insight that suggests the use of advanced tools from dynamical systems theory may help better understand gradient propagation.
\begin{figure}[!t]
\centering{}
\vspace*{-1mm}
\includegraphics[width= \linewidth]{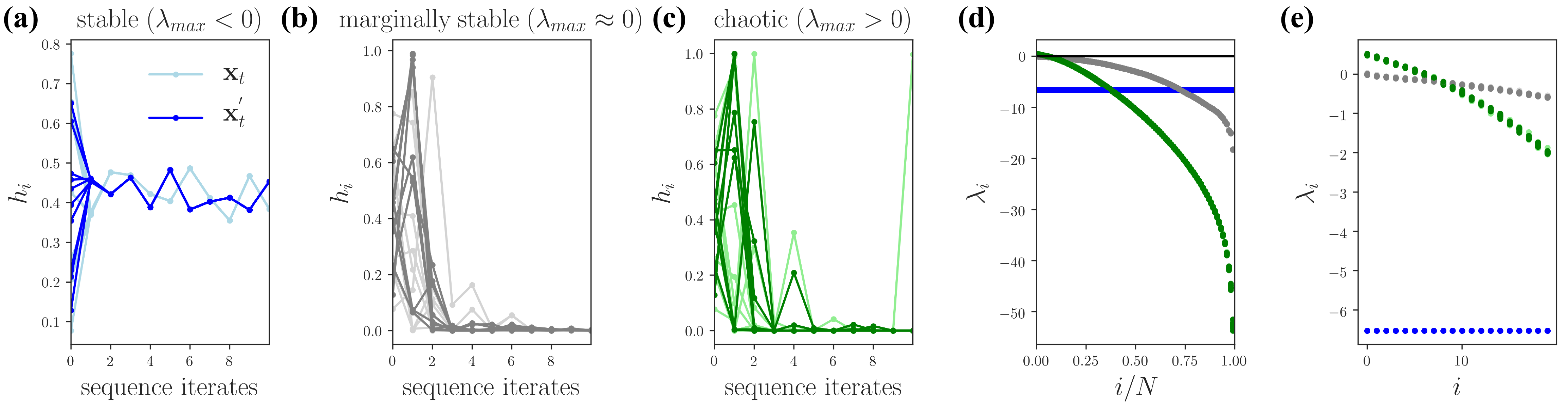}

\caption{{\it Distinguishing stability regimes of forward dynamics using Lyapunov spectra}. (a-c) Example unit activations from simulations computed from 10 initial conditions and 2 input sequences ($\mathbf{x}_t$ (light) and $\mathbf{x}_t'$ (dark)) for variance of weight matrix $\mathbf{V}$ set to gives 3 examples spanning the 3 qualitatively distinct stability regimes based on the sign of $\lambda_{max}$: stability, $\lambda_{max}<0$ (a; blue,  $\sigma^2_\mathbf{V}=1/500$); marginal stability, $\lambda_{max}\approx0$ (b; gray, $\sigma^2_\mathbf{V}=100$); and chaos $\lambda_{max}>0$ (c; green,  $\sigma^2_\mathbf{V}=500$). 
(d) Lyapunov spectra computed from (a-c) (same colors). (e) First 20 exponents from (d). (vanilla RNN with $\phi=\tanh$; $\mathbf{V}$ orthogonal;  $\mathbf{U}=\mathbb{I}$; 2 realizations (light/dark) of $\mathbf{x}$ Gaussian, $\mathbf{\Lambda}_\mathbf{x}=\sigma^2_\mathbf{x}\mathbb{I}$ with $\sigma^2_\mathbf{x}=0.6$;  10 Gaussian initial conditions, $\mathbf{\Lambda}_{\mathbf{h}_0}=\sigma^2_{\mathbf{h}_0}\mathbb{I}$ with $\sigma^2_{\mathbf{h}_0}=1$).}
\label{fig:untrained}
\vspace*{-2mm}
\end{figure}
\section{Background}
Here we describe the stochastic Lyapunov exponents from the ergodic theory of non-autonomous dynamical systems and outline their connection to the conditions that support gradient-based learning in RNNs. Jacobians are linear maps that evolve state perturbations forward along a trajectory, $\mathbf{u}_{t}=\mathbf{J}_{t}\mathbf{u}_{t-1}$, {\it e.g.} for some initial perturbation $\mathbf{h}_0(\epsilon)=\mathbf{h}_0+\epsilon \mathbf{u}_0$ with $\epsilon\ll 1$ and a given vector $\mathbf{u}_0$. Thus, perturbations can vanish or explode under the hidden state dynamics according to the singular value spectrum of $\mathbf{T}_t=\mathbf{J}_t\cdots \mathbf{J}_1$. The linear stability of the dynamics is obtained taking $\epsilon\to0$ and then $t\to\infty$. In particular, the Lyapunov exponents, $\{\lambda_i\}_{i=1}^N$, are the exponential growth rates associated with the singular values of $\mathbf{T}_t^{1/t}$ for $t\to\infty$, {\it i.e.} the logarithms of the eigenvalues of $\lim_{t\to\infty}\left(\mathbf{T}_t^\top\mathbf{T}_t\right)^\frac{1}{2t}$. As a property of the stationary dynamics, the Lyapunov exponents are independent of initial condition for ergodic systems, i.e. those with only one or the same type of attractor. If the maximum Lyapunov exponent $\lambda_{\textrm{max}}$ is positive, the stationary dynamics is chaotic and small perturbations explode, otherwise it is stable and small perturbations vanish. The theory was initially developed for autonomous dynamical systems, where stable dynamics implies limit-cycle or fixed-point attractors. How the shape of the Lyapunov spectrum varies with network model parameters has provided unprecedented insight into autonomous ( evolving in the absence of inputs) neural network models in theoretical neuroscience \cite{Monteforte2010,Engelken,PuelmaTouzel2015a}. 

RNN dynamics are non-autonomous because the hidden state dynamics $\mathbf{h}_t$ are driven by inputs $\mathbf{x}_t$. The theory of {\it random dynamical systems} generalizes stability analyses to non-autonomous dynamics driven by an input sequence sampled from a stationary distribution \cite{arnold2013random}. Analytical results typically employ uncorrelated Gaussian inputs, but the framework is expected to apply to a wider range of well-behaved input statistics. This includes those with finite, low-order moments and finite correlation times like character streams from written language and sensor data from motion capture systems. The time course of the driven dynamics depends on the specific input realization but, critically, the theory guarantees that the stationary dynamics for all input realizations share the same stability properties, which will in general depend on the input distribution (e.g., its variance). Stability here is quantified using the same definition of Lyapunov exponents in the autonomous case (now called {\it stochastic Lyapunov exponents}; we hereon will omit stochastic for brevity). In Fig.~\ref{fig:untrained}, we show vanilla RNN dynamics in stable, marginally stable, and chaotic regimes, as measured by the sign of the largest LE. For stable driven dynamics, the stationary activity on the time-dependent attractor (called a random sink) is independent of initial condition. This holds true also in the marginally stable case, $\lambda_{max}\approx0$, where in addition there are directions along which the sizes of perturbations (error gradients) are maintained over many iterates. For chaotic driven dynamics, the activity variance over initial conditions fluctuates in time in a complex way. Understanding stability properties in this non-autonomous setting is at the frontier of analysis of neural network dynamics in both theoretical neuroscience~\cite{Engelken,Lajoie2013,Hennequin2012} and machine learning~\cite{NIPS2019_9513,Liu}.

The practical calculation of the Lyapunov spectra (see Sec.~\ref{sec:praccalc}) is fast and offers more robust numerical behaviour and generality over singular value decomposition. Together, this suggests the Lyapunov spectra as a useful diagnostic for RNN sequence learning. 

There are a handful of features of the Lyapunov spectrum that dictate gradient propagation and thus, influence training speed and performance. The maximum LE determines the linear stability. The mean exponent determines the rate of contraction of full volume elements. The LE variance measures heterogeneity in stability across different directions and can reflect the conditioning of the product of many Jacobians. We use the first two in the analysis of our experiments later in the paper.

We remind readers that much of the theory regarding high-dimensional dynamics is derived in the so-called thermodynamic limit $N\to\infty$. Here, the number of exponents diverges and the spectrum over $i/N$ becomes stationary (so called extensive dynamics), i.e., it is insensitive to $N$ so long as $N$ is large enough (e.g. \cite{Monteforte2010,Engelken}). Here, self-averaging effects can take hold and enable accurate analytical results based on Gaussian assumptions justified by central limit theorem arguments. `Large enough' $N$ is typically in the hundreds (though this should be checked for any given application) and so these results can be useful for studying RNNs.

Finally, we note that obtaining the Lyapunov exponents from their definition relates to the singular values of $\mathbf{T}_t^{1/t}$ for large $t$, and thus is numerically impractical. The standard approach \cite{Benettin1980} is to exploit the fact that $m$-dimensional volume elements grow at a rate $\lambda^{(m)}=\sum_{i=1}^m\lambda_i$ and so the desired rates,  $\lambda_1=\lambda^{(1)}$, $\lambda_2=\lambda^{(1)}-\lambda_1$, ... arise from volumes obtained by projecting orthogonal to subspaces of increasing dimension. These rates are a direct output of computationally efficient orthonormalization procedures such as QR-decomposition. Next, we discuss the practical aspects of computing the Lyapunov exponents in data-driven RNNs.
\vspace*{-1mm}
\section{Lyapunov Spectrum Estimation for RNNs}\label{sec:praccalc}
\vspace*{-1mm}
We extend this framework of Lyapunov exponents to recurrent neural networks by calculating the asymptotic trajectories of the hidden states of the networks when driven by the same signal. 

\subsection{Algorithm description}
We adopt the well-established standard algorithm for computing the Lyapunov spectrum~\cite{Benettin1980,Dieci:1995gp}.
We choose the driving signal to be sampled from fixed-length sequences of the test set. 
For each input sequence in a batch, a matrix \textbf{Q} is initialized as the identity, and the hidden states, \textbf{h}, are initialized as zero. To track the expansion and contraction of \textbf{Q} at each step, the Jacobian of the hidden states is calculated and then applied to a set of vectors of \textbf{Q}. The expansion of each vector is calculated and the matrix \textbf{Q} is updated using the QR decomposition at each step. If $r_i^t$ is the expansion of the $i^{th}$ vector at time step $t$ -- corresponding to the $i^{th}$ diagonal element of \textbf{R} in the QR decomposition-- then the $i^{th}$ Lyapunov exponent $\lambda_i$ resulting from an input signal of length $T$ is given by:
\begin{equation}\label{eq:Lyap_sum}
    \lambda_i = \frac{1}{T}\sum_{t=1}^T {\log}(r_i^t)
\end{equation}

\begin{figure}[t!]
	\centering
\vspace*{-1mm}
	\begin{subfigure}{.4\linewidth}
	   \includegraphics[width= .8\linewidth]{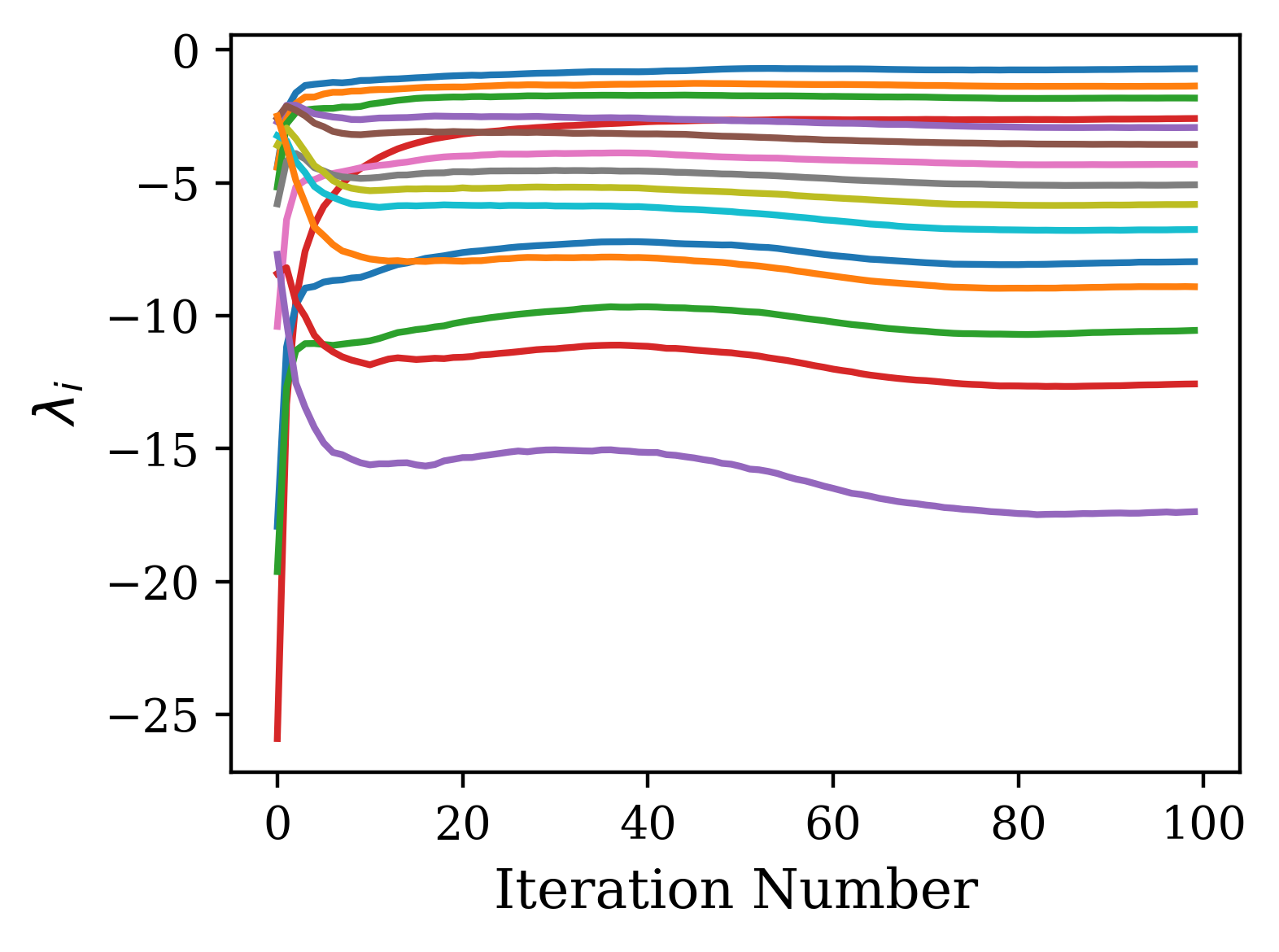}
        \caption{}
    \end{subfigure}
    \begin{subfigure}{.4\linewidth}
	   \includegraphics[width= .8\linewidth]{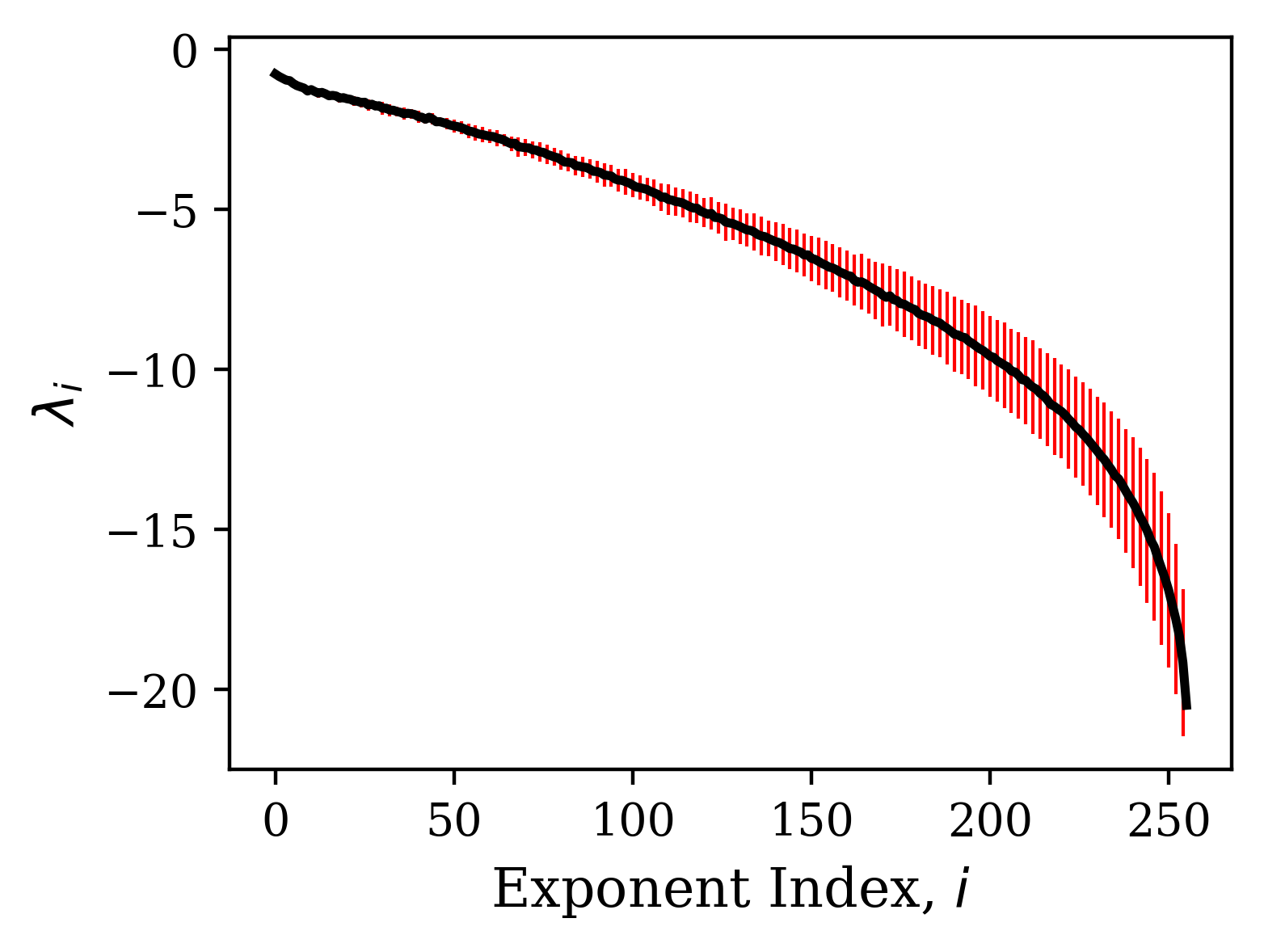}
	   	\caption{}
	\end{subfigure}
\caption{{\it Convergence of Lyapunov spectrum estimator}. (a) Lyapunov exponent estimation as a function of time, as shown in Equation \eqref{eq:Lyap_sum}, where $T$ is the number of iterations. Each line represents a different exponent in the spectrum. By 100 iterations, it is clear that each exponent is changing very little as a function of iteration number. (b) The mean Lyapunov spectrum over the set of input sequences. Standard deviation is shown by the red bars.}
\label{fig:Convergence}
\vspace*{-2mm}
\end{figure}

The Lyapunov exponents resulting from each input $x^j$ in the batch of input sequences are calculated in parallel and then averaged.
For our experiments, the Lyapunov exponents were calculated over 100 time steps with 10 different input sequences. The mean of the 10 resulting Lyapunov spectra is reported as the spectrum. An example calculation of the Lyapunov spectrum is shown in Fig.~\ref{fig:Convergence}.

The algorithm for computing the Lyapunov exponents is described in Algorithm \ref{algo: lyap}.

\begin{algorithm}[H]
    \SetAlgoLined
    \For{$\mathbf{x}^j$ in \text{Batch}}{
    initialize \textbf{h}, \textbf{Q}\\
    \For{t = 1 $\rightarrow$ T}{
    \textbf{h} $\leftarrow$ f($\textbf{h}$, $\mathbf{x}_t^j$)\\
    J $\leftarrow \frac{d\textbf{f}}{d\textbf{h}}$\\
    Q $\leftarrow$ J$\cdot$Q\\
    Q, R $\leftarrow$ $qr$(Q)\\
    $\gamma_i$ += log(R$_{ii}$)
    }
    $\lambda^j_i = \gamma^j_i/T$
    }
    $\lambda_i$ = mean$_j(\lambda_i^j)$
    \caption{Lyapunov Exponents Calculation}
    \label{algo: lyap}
\end{algorithm}

\vspace*{-1mm}
\section{Experiments}
\vspace*{-1mm}
To illustrate the computation and application of Lyapunov exponents, we applied the approach to two tasks with distinct task constraints: character prediction from sentences, where performance depends on capturing low-dimensional and long temporal correlations, and signal prediction from motion capture data, where performance relies also on signal correlations across the many dimensions of the input.

\subsection{Task details} \label{sec: tasks}
For character prediction, we use Leo Tolstoy's \textit{War and Peace} as the data and follow the character-level language modelling task outlined by Karpathy et al. in \cite{karpathy2015visualizing}. For signal prediction, we use the CMU motion-capture dataset and pre-process it using the procedure outlined in \cite{Li_2018_CVPR}. Both tasks require predicting the next step in a sequence given the preceding input sequence.
For the character prediction task (CharRNN), we use input sequences of 100 characters. For the motion capture task (CMU Mocap), we use input sequences of 25 time steps.

For each task, we use a single-layer LSTM architecture with weight parameters initialized uniformly on $[-p, p]$, where $p$ is referred to as the initialization parameter. For further details on the implementation of both tasks, see Supplementary Materials and in addition the Gihub repository\footnote{https://github.com/shlizee/lyapunov-hyperopt} for available code and ongoing work.

\subsection{Algorithm convergence properties}
\begin{figure}[!t]
\centering
\vspace*{-1mm}
	\begin{subfigure}{.19\linewidth}
	   \centering
	   \caption{}
	   \includegraphics[width= \linewidth]{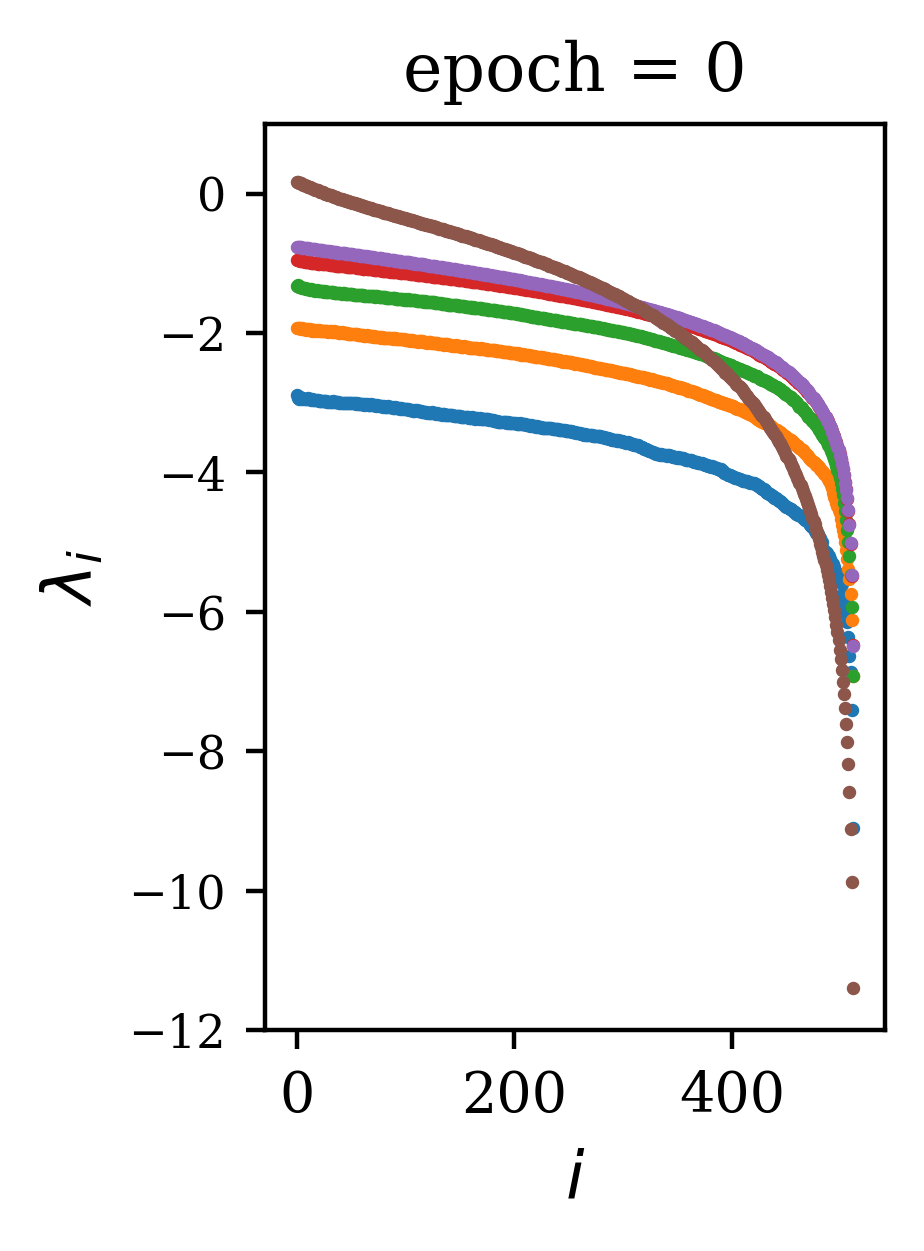}
    \end{subfigure} 
    \begin{subfigure}{.19\linewidth}
	   \centering
	   \caption{}
	   \includegraphics[width= \linewidth]{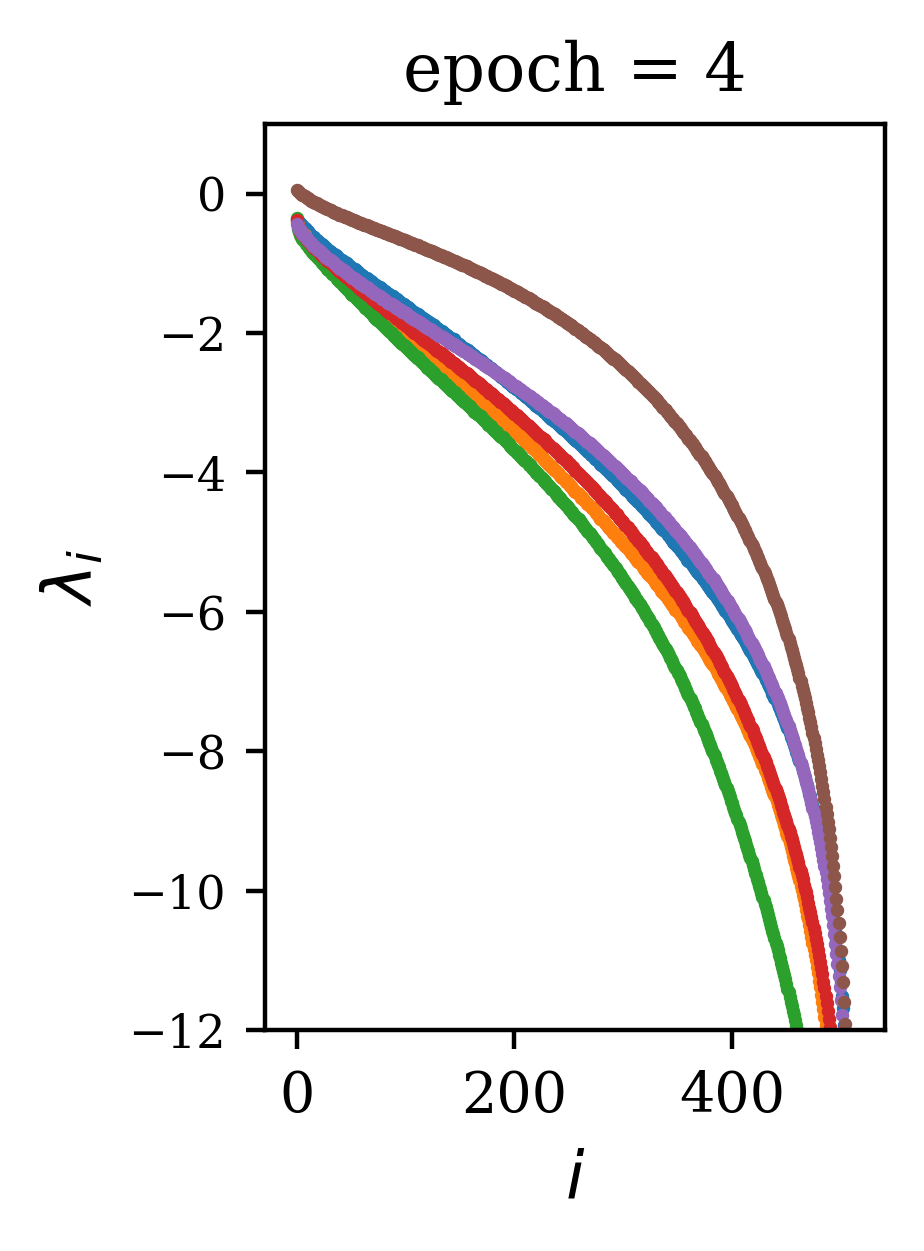}
	   
	\end{subfigure}
	\begin{subfigure}{.19\linewidth}
	   \centering
	   \caption{}
	   \includegraphics[width= \linewidth]{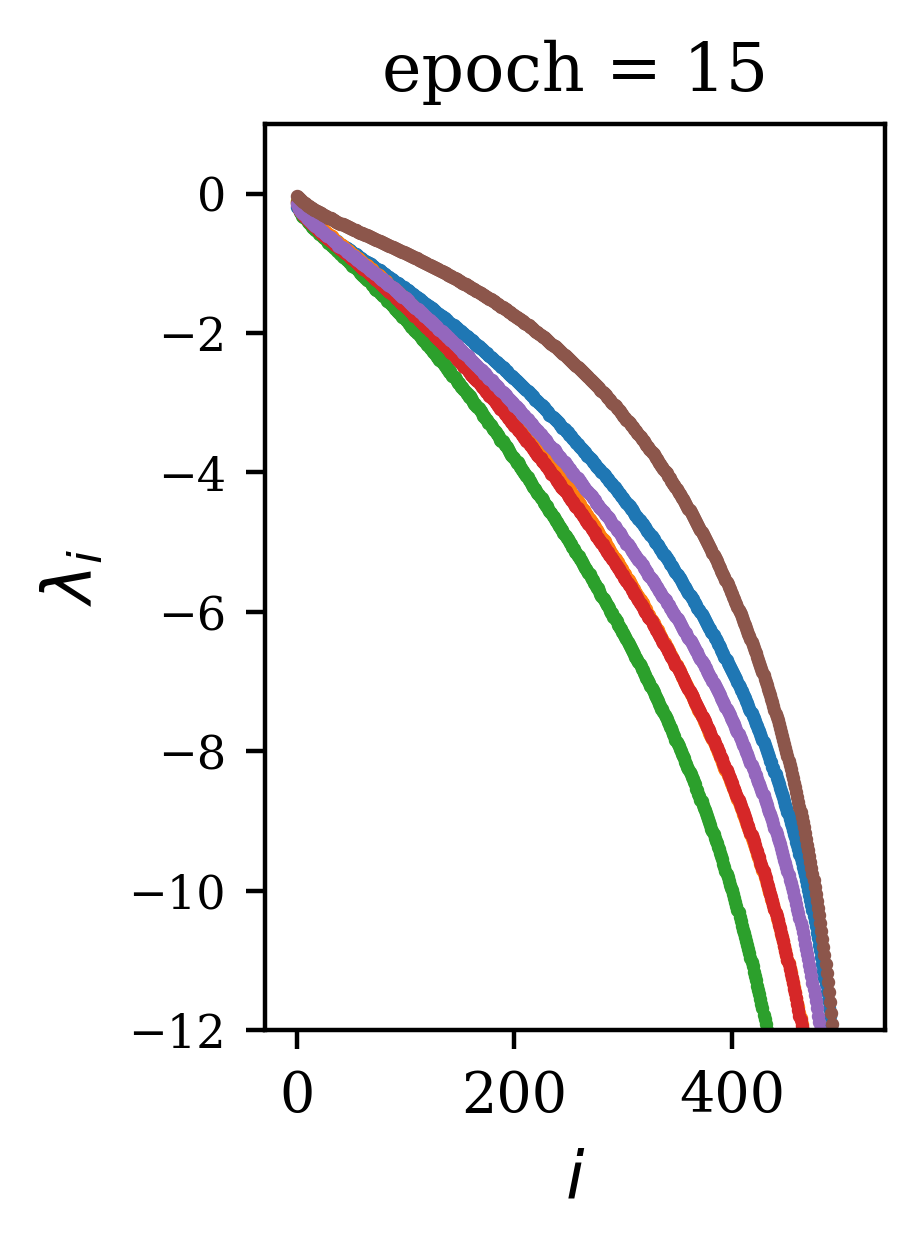}
	\end{subfigure}
	\begin{subfigure}{.19\linewidth}
	   \caption{}
	   \centering
	   \includegraphics[width= \linewidth]{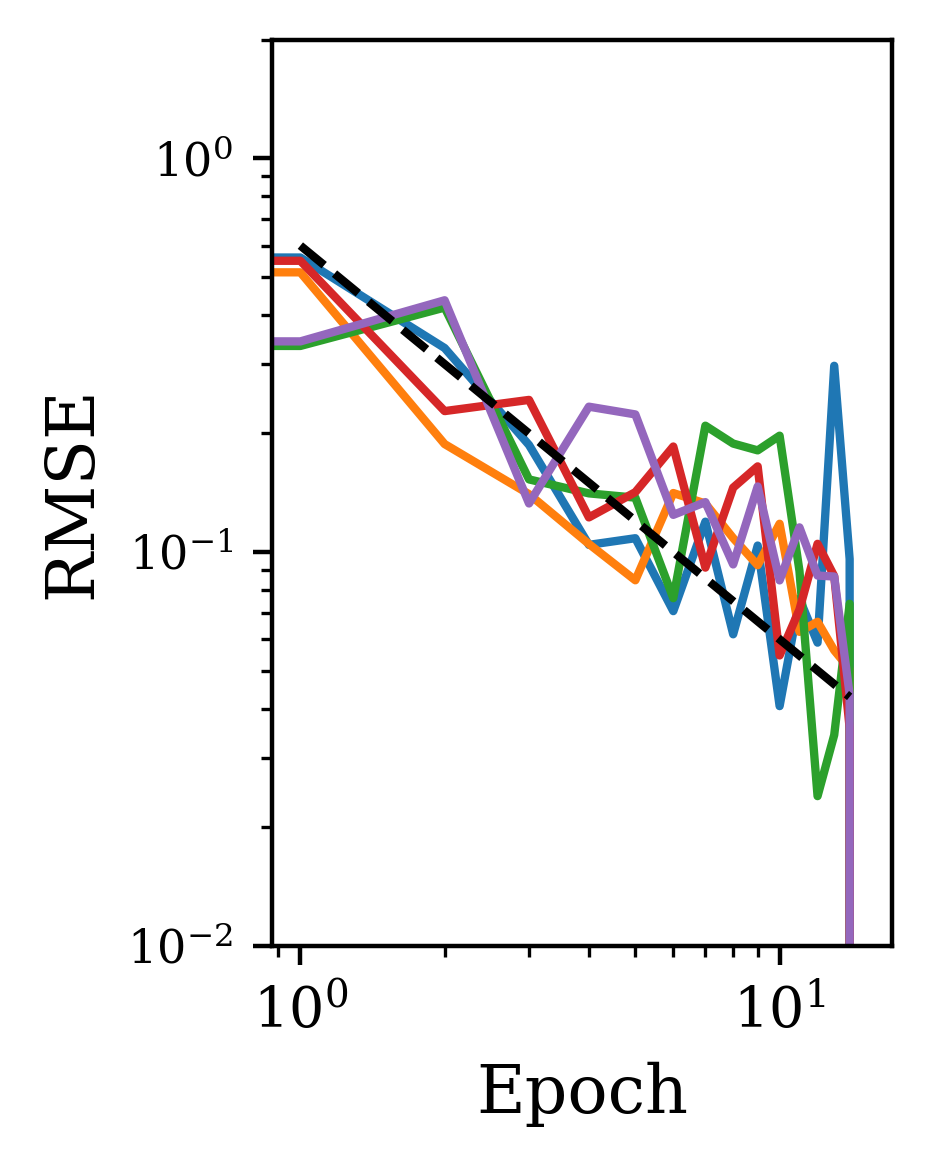}
	\end{subfigure}
	\begin{subfigure}{.19\linewidth}
	   \caption{}
	   \centering
	   \includegraphics[width= \linewidth]{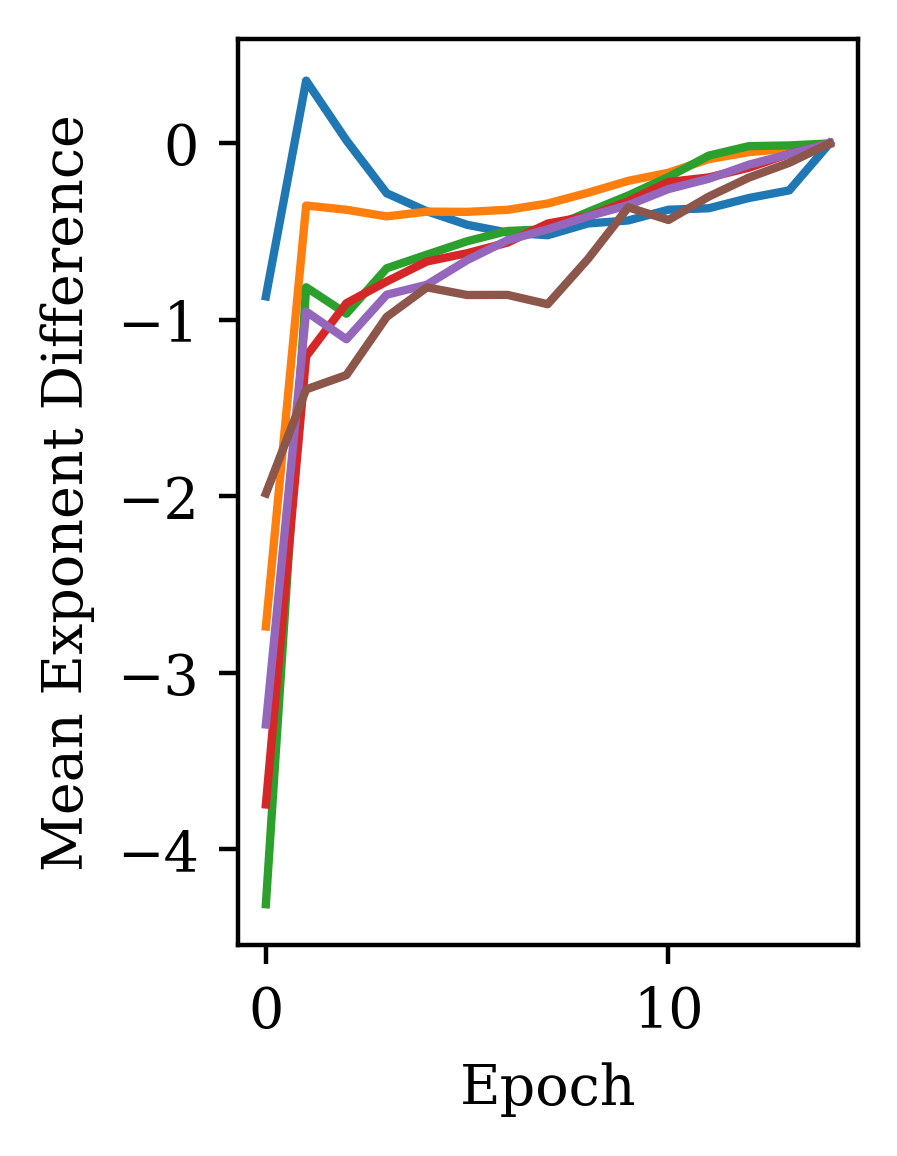}
	\end{subfigure}
\caption{{\it Evolution of the Lyapunov spectra over training}. For values of initialization parameter given in legend in (a) ($p = 0.04$ (Blue), $p = 0.08$ (Orange), $p=0.12$ (Green), $p=0.16$ (Red), $p=0.2$ (Purple), $p=0.5$ (Brown)). The spectra before training (a) and after four (b) and fifteen (c) epochs. The reduced mean-squared error between successive epochs (d) shows that the each spectrum converges quickly during training, with $t^{-1}$ plotted as a dashed black line for reference. Similarly, the difference between the mean of the Lyapunov exponents at the fifteenth epoch and the mean at previous epochs (e) rapidly approaches zero throughout training.}
\label{fig:Inits}
\vspace*{-1mm}
\end{figure}

We find that the general shape of the spectrum is reached early in training. In Fig. \ref{fig:Inits}, we use the CharRNN task as an example to show that the spectrum rapidly changes in the first few epochs of training, quickly converging after a small number of training epochs to a spectrum near that of the trained network. The reduced mean-squared difference shows a $\mathcal{O}(t^{-1})$ convergence with learning epoch $t$, while the mean difference shows this convergence is from below. This was true across the range of initialization parameters tested.
\subsection{Performance efficiency relative to training}
The calculation of the Lyapunov spectrum is very efficient relative to the computation required for training, as long as the user has an analytical expression for the Jacobian of the recurrent layer (as we describe the expressions for GRU and LSTM in supplementary materials). The average time required for calculating the spectrum for a network by evolving 10 input sequences for 100 time steps relative to the average time required to train those networks on the same device is shown in Table~\ref{table: speed} for the two tasks we consider (c.f. section~\ref{sec: tasks}). The mean number of training units (epochs or iterations) needed to reach the minimum validation loss is calculated for each task, and the total training time is compared to the Lyapunov spectra calculation time. For both tasks, the time required to calculate the spectrum was equivalent to about $4\%$ of the total training time.

\begin{table}[h!]
  \caption{{\it LE calculation rate relative to training}}
  \label{table: speed}
  \centering
  \begin{tabular}{lrrr}
    \toprule
    \multicolumn{1}{c}{} &Training Units& LE Calculation & Fraction of Total \\
    Example  & Needed & Time (training units) & \multicolumn{1}{c}{Training Time}\\
    \midrule
    CharRNN    & $26 \pm 10$ epochs & $1.04\pm .04$ epochs & \multicolumn{1}{c}{$0.04\pm 0.02$}  \\
    CMU Mocap     & $2.4 \pm .5 \times 10^3$ iters & $104\pm 3$ iters &  \multicolumn{1}{c}{$0.04\pm 0.01$} \\
    
    \bottomrule
\end{tabular}
\end{table}

Further improvements to the computation time of the exponents can be made by making slight modifications to Alg.~\ref{algo: lyap}. First, we can "warm-up" the hidden states \textbf{h} and \textbf{Q} onto the attractor of the dynamical system without the computational cost of the QR decomposition, allowing a faster convergence of the spectrum. A further improvement can be made by not taking the QR decomposition every time step during the calculation of the exponents, but instead every $T_{ON}$ steps. Since the QR step is the most expensive computation in the algorithm, the increase in speed over orthonormalizing every step is approximately $T_{ON}$. However, since increasing this interval leads to greater expansion and contraction of the vectors of \textbf{Q} before orthonormalization steps, this can lead to a spurious plateau at higher indices due to the accumulation of rounding errors. However, if one cares only about the first few exponents, this effect is negligible for reasonable selections of $T_{ON}$ (see supplementary materials for details about warmup and effect of increasing $T_{ON}$).

\vspace*{-3mm}
\subsection{Lyapunov spectrum as a robust readout of training stability}
In general, the dependence of the Lyapunov spectrum on hyperparameters is tangled. To direct our exploration of this dependence and how it relates to performance, we used the task constraints to guide our interpretation of spectra behavior of trained networks from randomly sampled hyperparameters.

\begin{figure}[t!]
\vspace*{-1mm}
	\begin{subfigure}{.49\linewidth}
	   \centering
	   \includegraphics[width= \linewidth]{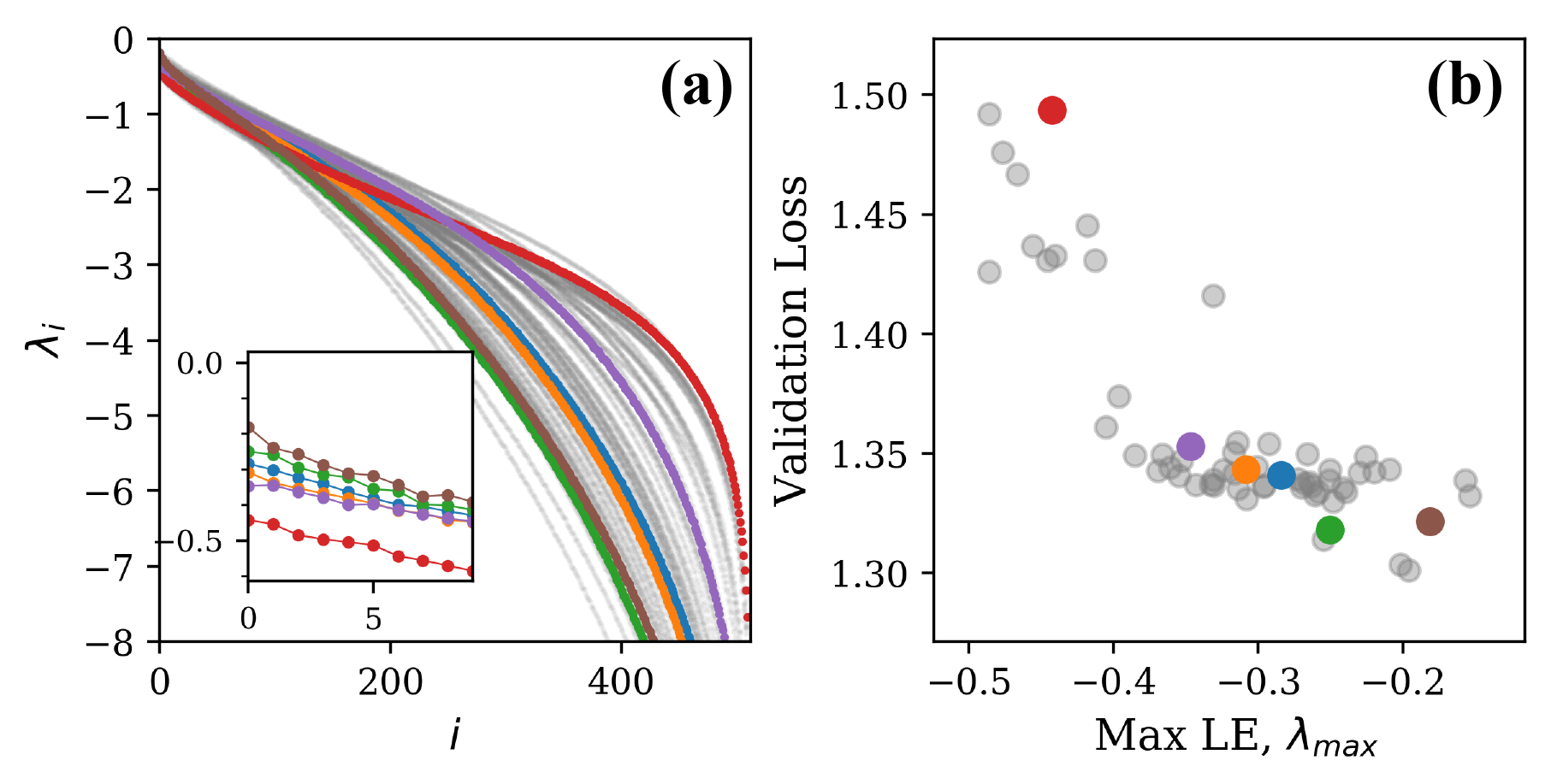}
    \end{subfigure}
    \begin{subfigure}{.49\linewidth}
	   \centering
	   \includegraphics[width= \linewidth]{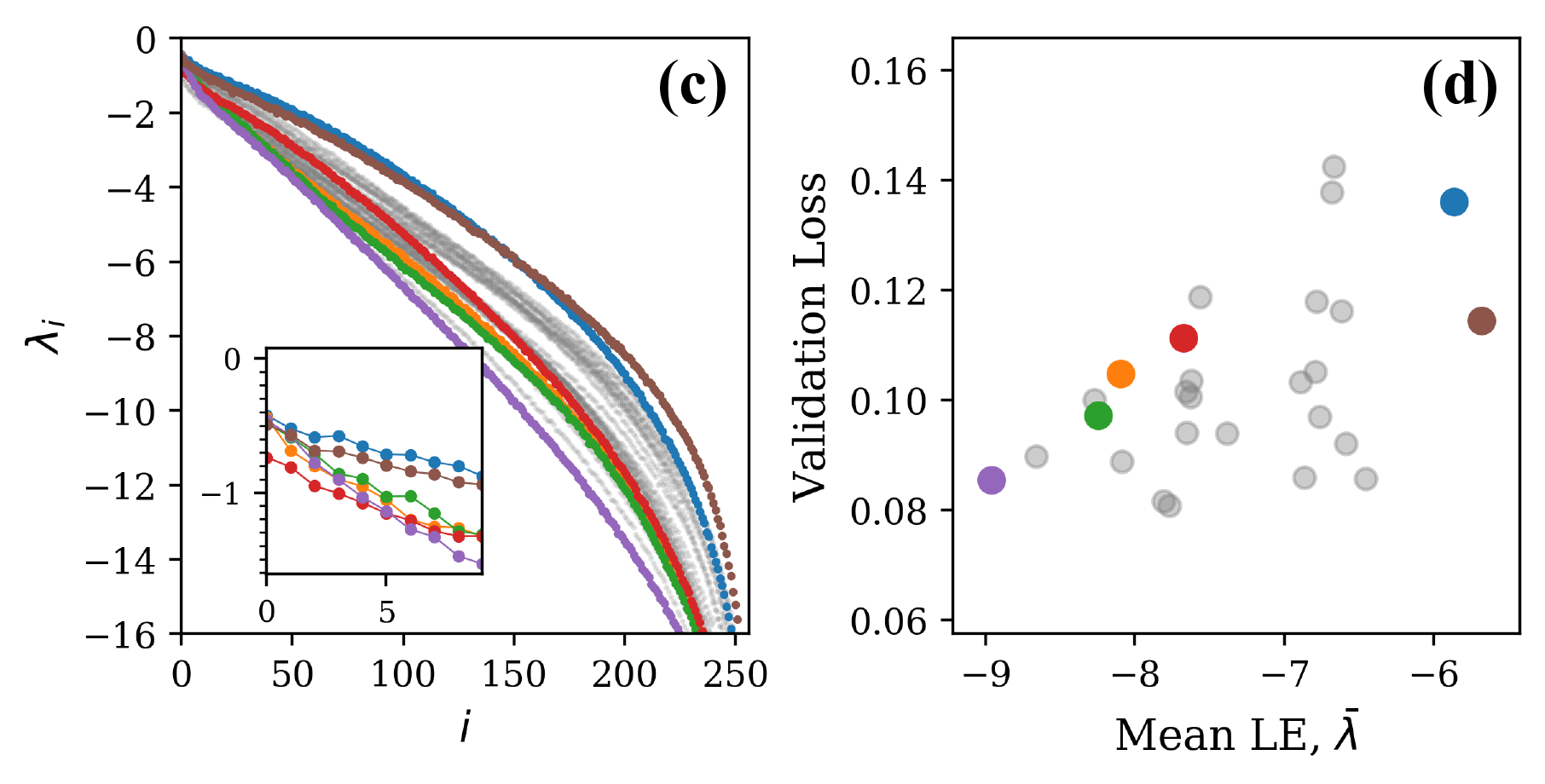}
	\end{subfigure}
\caption{{\it Correlation between Lyapunov spectra and performance}. Shown are spectra (a,c) and loss dependence on spectra (b,d) for networks trained on CharRNN (a,b) and CMU MoCap (c,d). Those in (a,b) are for 60 random combinations of learning rate and log-dropout rate, while those in (c,d) are for 30 values of the initialization parameter, $p$. The colors in (a,b) and (c,d) are 6 samples selected by eye to roughly span the range of gross spectra shapes seen in (a) and (c), respectively. All other spectra are shown in gray.
}
\label{fig:performance}
\vspace*{-3mm}
\end{figure}

For CharRNN, we hypothesized from existing work that to satisfy the constraint of propagating a scalar signal over many iterates, it would be sufficient to have a single exponent approaching 0 with the other exponents more negative~\cite{Henaff2015}. Focusing here on hyperparameters, we uniformly sampled a fixed range of log-dropout and learning rate while keeping the initialization parameter fixed. In Fig.~\ref{fig:performance}(a,b), we observe the spectra and indeed find a correlation between maximum Lyapunov exponent and validation loss. Networks with a maximum LE closer to zero indeed performed better. Of these 60 spectra, we selected a smaller subset of 6 that spanned the range of the gross shapes. Interestingly, they also spanned the range of losses, suggesting there is a consistent signal about the loss encoded in the complex, yet systematic variation of the maximum Lyapunov exponent with hyperparameters. The gross shape of these spectra correlated less with the loss, supporting our hypothesis. A straightforward interpretation of the signal we have observed is that the slower the activity in the mode associated with this maximum LE, the longer gradient information can propagate backward in time. This would give the dynamics more predictive power and thus better performance, but it remains to be verified.
Next, we focused on the motion capture prediction task. Here, the higher dimensional data means that the changes that training induces in the shape of the Lyapunov spectra will likely be more complicated. We hypothesized that nevertheless there would be more information in the gross shape of the spectra of the dynamics of trained networks and less in the maximum LE. Here, we vary the initialization parameter from 0.05 to 0.5 while keeping the dropout and learning rates fixed. Indeed, we find (Fig.~\ref{fig:performance}c,d) consistent variation of the gross shape with loss. We have quantified the variation of loss with gross shape using the mean Lyapunov exponent, though the spectra vary with higher order features as well.

The plots of Fig.~\ref{fig:performance} illustrate that distinct properties of the LE spectrum correlate with validation performance, depending upon task structure.
For the CharRNN task, the maximum exponent being larger (closer to 0), corresponds to better performance. For the Mocap learning task, the larger mean Lyapunov exponent correspond to better performance. 
Please see  Supplementary Materials for other spectrum statistics of all networks tested.

Given our observation that the spectrum changes most rapidly early in training, we believe the most strongly-correlated metrics for a task could allow an alternative way to asses performance of a network early in training, thus serving as a hyper-parameter tuning tool.

\vspace*{-1mm}
\section{Conclusion \& Discussion}
\vspace*{-1mm}
In this paper, we presented an exposition and example application of Lyapunov exponents for understanding training stability in RNNs. We motivated them as a natural quantity related to stability of dynamics and useful in a natural complementary approach to existing mathematical approaches for understanding training stability focused on the singular value spectrum. We adapted the standard algorithm for their computation to RNNs, and showed how it could be made more efficient in standard machine learning development environments. We demonstrated that even the basic implementation runs almost 2 orders of magnitude faster than typical training time (see~\cite{Engelken} for further precision and efficiency considerations) and that it converges relatively quickly with learning time. The latter implies it could be useful as a readout of performance early in training. To test this application, we studied it in two tasks with distinct constraints. In both we find interpretable variation with performance reflecting these distinct constraints. 

There are a few points to be made about LEs in RNNs. First, one should consider carefully the degrees of freedom in the architecture under consideration. We have considered the hidden state dynamics of an LSTM such that there are as many Lyapunov exponents as there are units. LSTMs have gating variables that can also carry their own dynamics and could be considered as degrees of freedom, adding another $N$ exponents to the spectrum. While we were able to obtain evidence from the spectra of hidden states alone, recent work \cite{Can2020,Engelken} suggests that studying the stability in the subspace of gating variables can also be informative. A second point is that one should also consider the role of the input weights $\mathbf{U}$ (c.f. Eq. \ref{eq:vanilla}). The strength of inputs (e.g. input signal variance) that drive high-dimensional dynamics has long been known to have a stabilizing effect~\cite{Molgedey1992,Lajoie2013,Schuecker2018}. Thus, the input statistics of the task's data can play a role that is modulated by $\mathbf{U}$, making the latter a target for gradient-based algorithms aiming to decrease loss. Due to limited space, we have not analyzed these weights here, but believe it is important to do so in applications.
Finally, we raise perhaps the most glaring adaptation: that of approximating this asymptotic quantity to settings of finite-length sequences. We proposed to use averaging over inputs, in addition to averaging over initial conditions, as a way to take advantage of batch tensor mechanics to achieve faster LE estimates. This is in contrast to long-time averages used in the LE definition, although our method is justified under assumptions of ergodicity and short transient dynamics. 
Analyzing the influence of input batch size on this precision is a topic of future research.

The task analysis we have performed suggests a use case for the Lyapunov spectrum as having features that serve as an early readout of performance useful to speed up hyperparameter search. We looked the maximum and mean Lyapunov exponent as two features highlighted by hypotheses we made based on our knowledge of the task constraints. Presumably, there are others relevant here and in other tasks. The search for generic features that do not require knowledge of task constraints (e.g. dynamical isometry) is important, as is demonstrating how useful these features are in practise with complex sequence data. One limitation is how quickly they converge with learning. For example, as a composite quantity, the mean LE converges rather quickly, while the maximum LE is significantly slower to converge (see supplemental material). In \cite{Engelken}, it is shown that loss function/learning rule combinations can also sculpt the spectra and alter its statistics, demonstrating that these are more sources of variation to understand. Interestingly, Full Force~\cite{Depasquale2018}, an alternative to Back-propagation-through-time, has a qualitatively similar spectra (same max, min, and mean LE) but with much lower variance, presumably a desirable feature for generalization. This method also demands more information about performance than just a gradient, suggesting the interesting hypothesis that the precision with which the features of the Lyapunov spectra can be sculpted can scale with the amount of information provided in training. Finally, recent work has given theoretical grounding to why LSTMs avoid vanishing gradients~\cite{Can2020}. Of course this was the reason for the design of LSTM so it only recapitulates the original design intuition. Looking forward however as more complex architectures are developed with less interpretable design, this approach based on LEs can still provide novel insight into why they work.

We close with a short discussion of open problems as this is a prominent area in understanding network dynamics at the intersection of machine learning and computational neuroscience. Having supporting analytical results is essential for robust control of complex problems. To this end, understanding how and where the assumption of untied weights breaks down and how forward propagation differs from backward propagation will be important in extending analytical results on spectral constraints. The Lyapunov spectrum can guide this work. Last, we believe insightful parallels in theoretical neuroscience work. For example, Lyapunov spectrum have been derived for additional degrees of freedom in the unit dynamics, different connectivity ensembles, and more. Also, a discrepancy between the loss of linear stability and the onset of chaotic dynamics in driven systems must be understood~\cite{Schuecker2018}. Making these connections explicit will serve both fields. 

\medskip

\bibliographystyle{unsrt}
\small

\appendix
\section{Task Implementation Details}
\subsection{CharRNN: Character Prediction Task}
For our CharRNN experiment, the text used for training the networks was Leo Tolstoy's \textit{War and Peace}, which contains about 3.2 million characters of mostly English text, with a total of 82 distinct characters in the vocabulary. We split the data into train, validation, and test sets with 80/10/10 ratio, respectively. We study the performance of single-layer LSTM networks (with no bias vectors) with hidden size of 512 units trained on this data set. We train the network using stochastic gradient descent with a batch size of 100 and the Adam optimizer. The network is unrolled for 100 time steps. After 10 epochs of training, the learning rate is decreased by a multiplying by a factor a .95 after each additional epoch. The batch size used for training was 100.

For each combination of dropout and learning rate, the network was trained for 40 epochs, and the training epoch at which the network had the lowest validation loss was selected.

\subsection{CMU Mocap: Signal Prediction Task}
The CMU Motion Capture (CMU Mocap) dataset is a high-dimensional data set capturing the location of various sensors on a subject over time as they conduct a variety of tasks. In particular, we narrow our focus onto the Sports and Activities category of the data set. The basketball, soccer, walking, and jumping subcategories were selected to compile a variety of motions which require consistent full-body movement (unlike, for example, the "directing traffic" and "car washing" sub-categories). For this task, we use a standard 256-unit LSTM.

We chose the training and validation split for the data corresponding to each task as 80/20. After this split, the data for all tasks considered was mixed into a single training and validation set. For each dimension of the data, the mean and the standard deviation across all times points is calculated, and those dimensions for which the standard deviation is less than $10^{-4}$ are ignored. All other dimensions are normalized by their standard deviation. For our example, we consider the first 20 features of the remaining data. The network is trained to predict the 20-dimensional output at the time step $t$ given the preceding input sequence. The loss function is defined as mean-squared error across the 20 dimensional-output. We chose the batch size  of 250 for training.

\subsection{Trained Network Hyperparameters}
The following are the network parameters for the networks highlighted for each task in Figure 4.
\begin{table}[h!]
  \caption{{\it CharRNN Networks Hyperparameters}}
  \label{table: CharRNN_Params}
  \centering
  \begin{tabular}{llll}
    \toprule
            &           & Learning  &Initialization \\
     Color  &Dropout    & Rate      &Parameter, \textit{p}\\
    \midrule
    \textcolor{blue}{Blue}    & 0.034 & 0.00301 & 0.08\\
    \textcolor{orange}{Orange}  & 0.000 & 0.00319 & 0.08\\
    \textcolor{green}{Green}   & 0.039 & 0.00574 & 0.08\\
    \textcolor{red}{Red}     & 0.028 & 0.00055 & 0.08\\
    \textcolor{purple}{Purple}  & 0.014 & 0.00153 & 0.08\\
    \textcolor{brown}{Brown}   & 0.106 & 0.00485 & 0.08\\
    \bottomrule
\end{tabular}
\end{table}

\begin{table}[h!]
  \caption{{\it CMU Mocap Networks Hyperparameters}}
  \label{table: Mocap_params}
  \centering
  \begin{tabular}{llll}
    \toprule
            &           & Learning  &Initialization \\
     Color  &Dropout    & Rate      &Parameter, \textit{p}\\
    \midrule
    \textcolor{blue}{Blue}    & 0.0 & 0.05 & 0.476\\
    \textcolor{orange}{Orange}  & 0.0 & 0.05 & 0.305\\
    \textcolor{green}{Green}   & 0.0 & 0.05 & 0.196\\
    \textcolor{red}{Red}     & 0.0 & 0.05 & 0.309\\
    \textcolor{purple}{Purple}  & 0.0 & 0.05 & 0.059\\
    \textcolor{brown}{Brown}   & 0.0 & 0.05 & 0.439\\
    \bottomrule
\end{tabular}
\end{table}

\section{Analytical Jacobian Expressions}
 We aim to calculate the derivative of the hidden states $h$ of the recurrent neural network at time $t$ with respect to the hidden states at the previous time step, $t-1$,
 $$ \frac{\partial h_t}{\partial h_{t-1}}.$$
 When the network has multiple recurrent layers, we can partition the hidden state $h$ the hidden states at layer $k$. Then, if a given recurrent network has $l$ layers,
 $$
 h_t = \left[\begin{matrix}
 h_t^1\\ h_t^2\\ \vdots \\h_t^k\\ \vdots\\ h_t^l
 \end{matrix}\right]
 $$
 We will denote the input of layer $k$ as $x^k_t$. We can easily use the equations for a given recurrent network type to calculate the derivative of the hidden state of a layer $k$, $h_t^k$ with respect to the hidden states of that same layer at the previous time step $h_t^k$, as well as with respect to the external input into that layer $x^k_t$. The expressions for $\frac{\partial h^k_t}{\partial h^k_{t-1}}$ and $\frac{\partial h^k_t}{\partial x^k_t}$ for LSTMs and GRUs are given in the following sections.\\
 Since layer $k$ takes the output of layer $k-1$ as input, $x^k_t = h_t^{k-1}$ for $k \geq 2$. Then, in order to find the derivative of the $k^{th}$ layer with respect to the a previous layer,
$$
\frac{\partial h^k_t}{\partial h^j_{t-1}}, \quad j\leq k ,
$$

we need to compose the derivative with the derivative with respect to the input for each preceding layer until layer $j$ is reached, at which point we calculate the derivative with respect to the hidden state at the previous time step, $\frac{\partial h^j_t}{\partial h^j_{t-1}}$. This gives the composition:
\begin{equation*}
    \frac{\partial h^k_t}{\partial h^{j}_{t-1}} = \frac{\partial h^k_t}{\partial h^{k-1}_t}\cdot \frac{\partial h^{k-1}_t}{\partial h^{k-2}_t} \cdot ... \cdot \frac{\partial h^{j+1}_t}{\partial h^j_t}\cdot\frac{\partial h^j_t}{\partial h^j_{t-1}}
\end{equation*}
\raggedright
For each $\frac{\partial h^i_t}{\partial h^{i-1}_t}, j<i\leq k$, we use the expression for $\frac{\partial h^i_t}{\partial x^i_t}$.

When implementing this in matrix form, we start by calculating the derivatives for the first layer, then each layer below it. The composition described above can be taken by multiplying by the block directly above it. This reduces redundant computation and requires only a loop as long as the number of layers in the RNN. Given the structure of one-directional recurrent layers, this matrix would be block lower-triangular, with the size of each block corresponding to the number of hidden units in the layer.

 \subsection{LSTM Derivatives}
The LSTM is defined by the equations
\begin{align*}
&f_t = \sigma_g\left(W_fx_t+U_fh_{t-1}+b_f\right)\\
&i_t = \sigma_g\left(W_ix_t + U_ih_{t-1} + b_i\right)\\
&o_t = \sigma_g\left(W_ox_t + U_oh_{t-1} + b_o\right)\\
&c_t = \sigma_h\left(W_cx_t + U_ch_{t-1} + b_c\right),
\end{align*}
where $\sigma_g$ is the sigmoid function  and $\sigma_h$ is $tanh$ (both applied element-wise).
We use the following notation when calculating these derivatives:
\begin{equation*}y_* = W_*x_t + U_*h_{t-1} + b_*,\end{equation*}
where $_*$ is determined by the gate/state. For LSTM, there are 4 different gates/states, ($f, ~i,~ o, ~c$), each with a corresponding $y_*, W_*, U_*, b_*$. We also use $\circ$ to indicate elementwise multiplication (if both vectors) or row-wise multiplication (if vector and matrix).\\
In other words, if $\textbf{v}, \textbf{u} \in \mathcal{R}^n, \textbf{W} \in \mathcal{R}^{n \times n}$, then
\begin{align*}
    \textbf{u} \circ \textbf{v} = \left[\begin{matrix}
 u_1 \cdot v_1\\ \vdots \\u_k \cdot v_k\\ \vdots\\ u_n \cdot v_n
 \end{matrix}\right], 
 \quad \textbf{u} \circ \textbf{W} =  \left[\begin{matrix}
 u_1 \cdot W_{11} &\hdots &  u_1 \cdot W_{1j} &\hdots &  u_1 \cdot W_{1n}\\
 \vdots && \vdots&&\vdots
 \\u_k \cdot W_{k1} &\hdots &  u_k \cdot W_{kj} &\hdots &  u_k \cdot W_{kn} \\
 \vdots && \vdots&&\vdots\\ 
 u_n \cdot W_{n1} &\hdots &  u_n \cdot W_{nj} &\hdots &  u_n \cdot W_{nn}\\
 \end{matrix}\right]
\end{align*}

The derivatives of each of these gates/states with respect to the hidden states and the input is shown below. The final line shows what the derivative of the hidden state is relative to the hidden state at the previous time step.
\begin{align*}
&\frac{\partial f_t}{\partial h_{t-1}} = \left[\sigma (y_f)\circ\left(1 - \sigma(y_f)\right)\right]^T \circ U_f\\
&\frac{\partial i_t}{\partial h_{t-1}} = \left[\sigma (y_i)\circ\left(1 - \sigma(y_i)\right)\right]^T \circ U_i\\
&\frac{\partial o_t}{\partial h_{t-1}} = \left[\sigma (y_o)\circ\left(1 - \sigma(y_o)\right)\right]^T \circ U_o\\
&\frac{\partial c_t}{\partial h_{t-1}} = \frac{\partial f_t}{\partial h_{t-1}}\circ c_{t-1} + \frac{\partial i_t}{\partial h_{t-1}}\circ \text{tanh}(y_c) + i_t\circ \text{sech}^2(y_c)\circ U_c\\
&\frac{\partial h_t}{\partial h_{t-1}} = \frac{\partial o_t}{\partial h_{t-1}}\circ \text{tanh}(c_t) + o_t \circ \text{sech}^2(c_t)\circ \frac{\partial c_t}{\partial h_{t-1}}
\end{align*}

\begin{align*}
&\frac{\partial f_t}{\partial x_t} = \left[\sigma (y_f)\circ\left(1 - \sigma(y_f)\right)\right]^T \circ W_f\\
&\frac{\partial i_t}{\partial x_t} = \left[\sigma (y_i)\circ\left(1 - \sigma(y_i)\right)\right]^T \circ W_i\\
&\frac{\partial o_t}{\partial x_t} = \left[\sigma (y_o)\circ\left(1 - \sigma(y_o)\right)\right]^T \circ W_o\\
&\frac{\partial c_t}{\partial x_t} = \frac{\partial f_t}{\partial x_t}\circ c_{t-1} + \frac{\partial i_t}{\partial x_t}\circ \text{tanh}(y_c) + i_t\circ \text{sech}^2(y_c)\circ W_c\\
&\frac{\partial h_t}{\partial x_t} = \frac{\partial o_t}{\partial x_t}\circ \text{tanh}(c_t) + o_t \circ \text{sech}^2(c_t)\circ \frac{\partial c_t}{\partial x_t}
\end{align*}

\section{LE calculation alternatives and their effects}
The algorithm for calculation of LE can be altered by introducing two alterations. ({\it c.f.} Algo. \ref{algo: lyap_mod}). 
First, we can `warm-up' the hidden states \textbf{h} and the vectors of \textbf{Q} by evolving them via the network dynamics for a set number of steps before calculating the Lyapunov Exponents. This allows the hidden states and vectors of \textbf{Q} to relax onto the attractor of the dynamical system defined by the network. This means that all samples used are from the stationary distribution on the attractor. Without the warmup, the initial samples are from a transient phase with a distribution over initial conditions that lies off the attractor and so has a bias effect on the average for finite sample averaging.

\begin{figure}[h!]
\vspace*{-2mm}
	\begin{subfigure}{.32\linewidth}
	   \caption{}
	   \centering
	   \includegraphics[width= \linewidth]{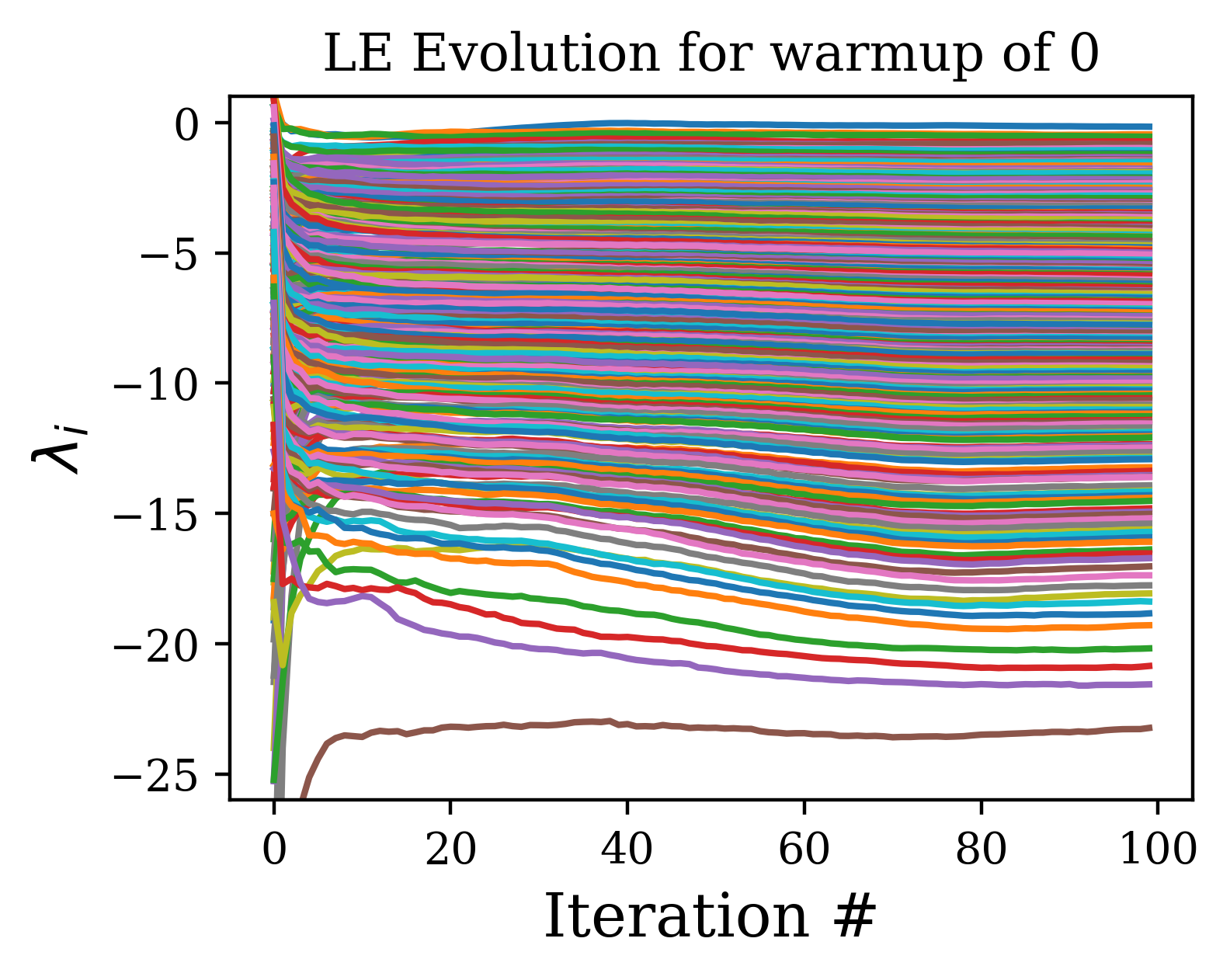}
    \end{subfigure}
    \begin{subfigure}{.32\linewidth}
	   \centering
	   \caption{}
	   \includegraphics[width= \linewidth]{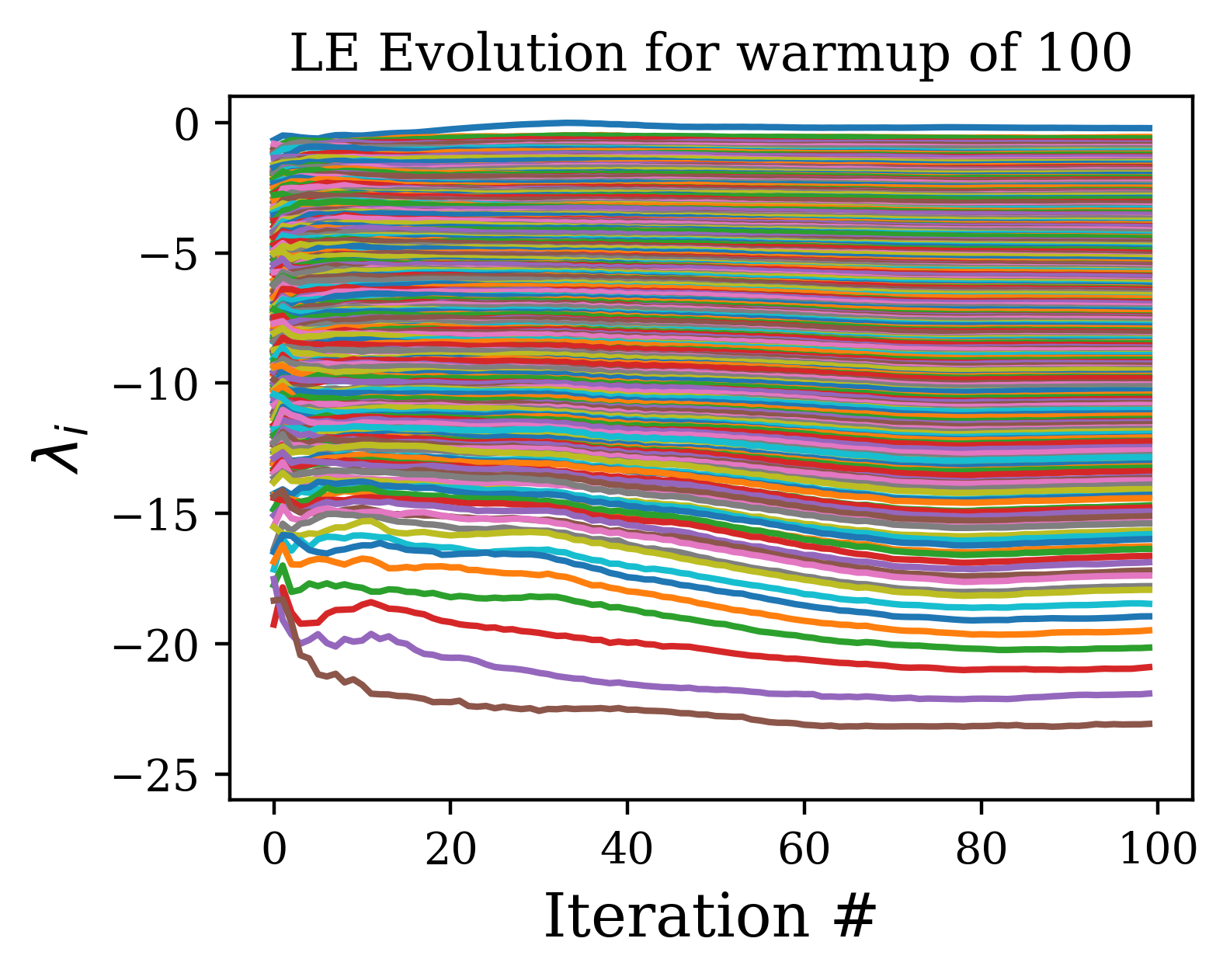}
	\end{subfigure}
	\begin{subfigure}{.32\linewidth}
	   \centering
	   \caption{}
	   \includegraphics[width= \linewidth]{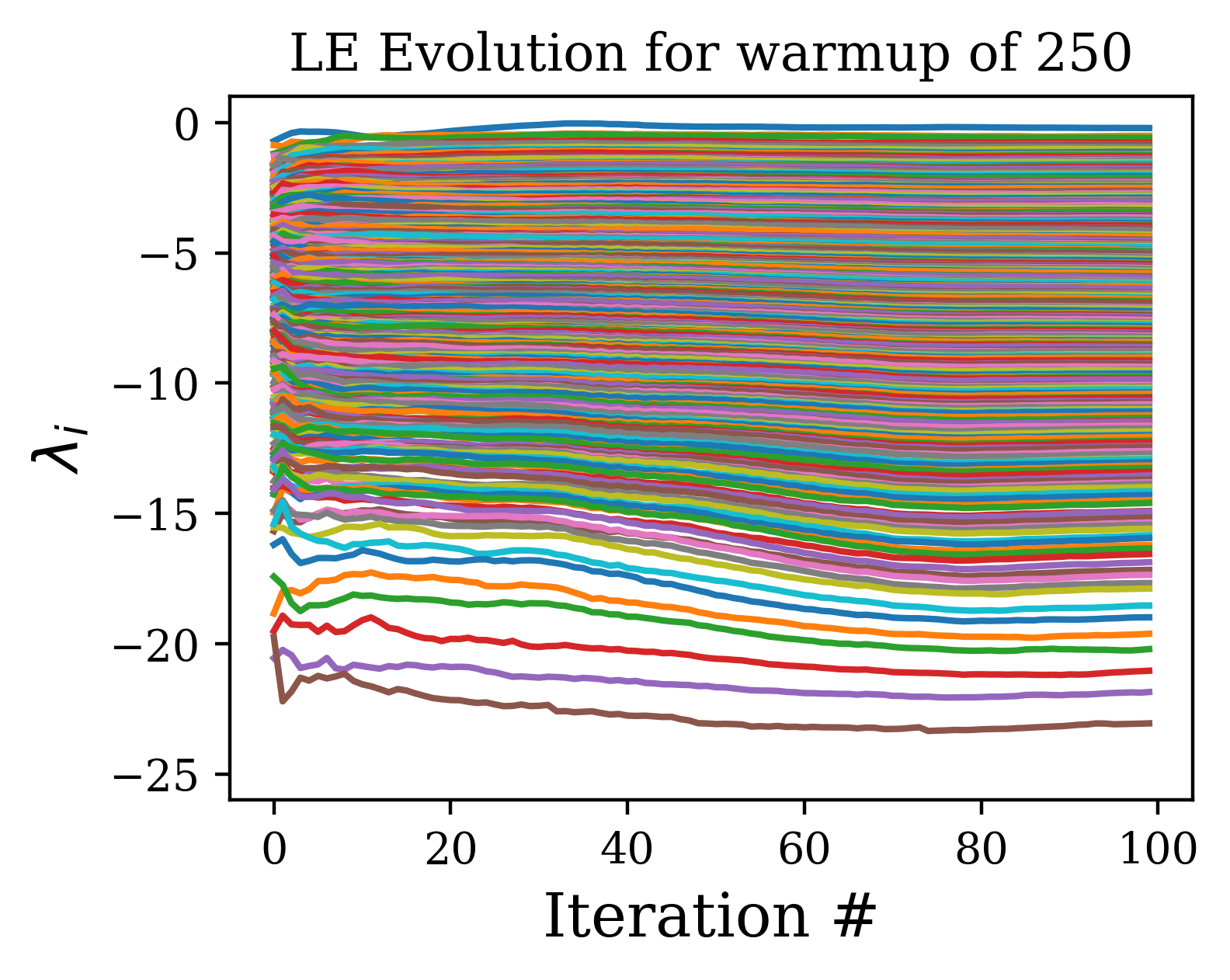}
	\end{subfigure}
\caption{{\it The effect of warmup on the calculation Lyapunov spectrum}. Shown are the Lyanpunov exponents as a function of iteration with (a) no warmup and warmups of (b) 100 time steps and (c) 250 time steps. It is clear that the case with no warmup has large changes early in the evolution of the spectrum, whereas those with warmup periods have much smaller changes early in the evolution.
}
\label{fig:warmup}
\vspace{-3mm}
\end{figure}

\begin{figure}[t!]
\vspace*{-1mm}
	\begin{subfigure}{.49\linewidth}
	   \caption{}
	   \centering
	   \includegraphics[width= \linewidth]{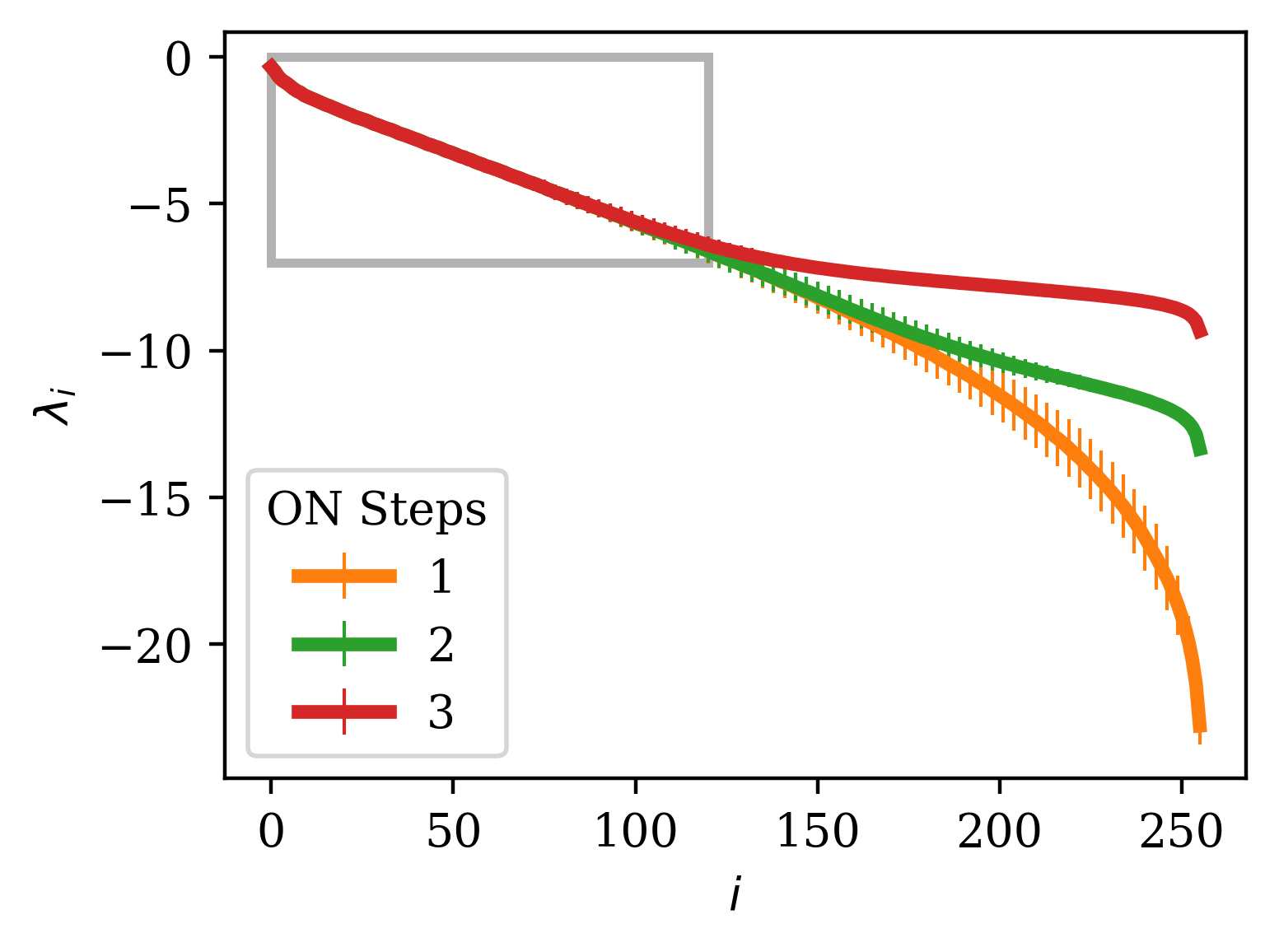}
    \end{subfigure}
    \begin{subfigure}{.49\linewidth}
	   \centering
	   \caption{}
	   \includegraphics[width= \linewidth]{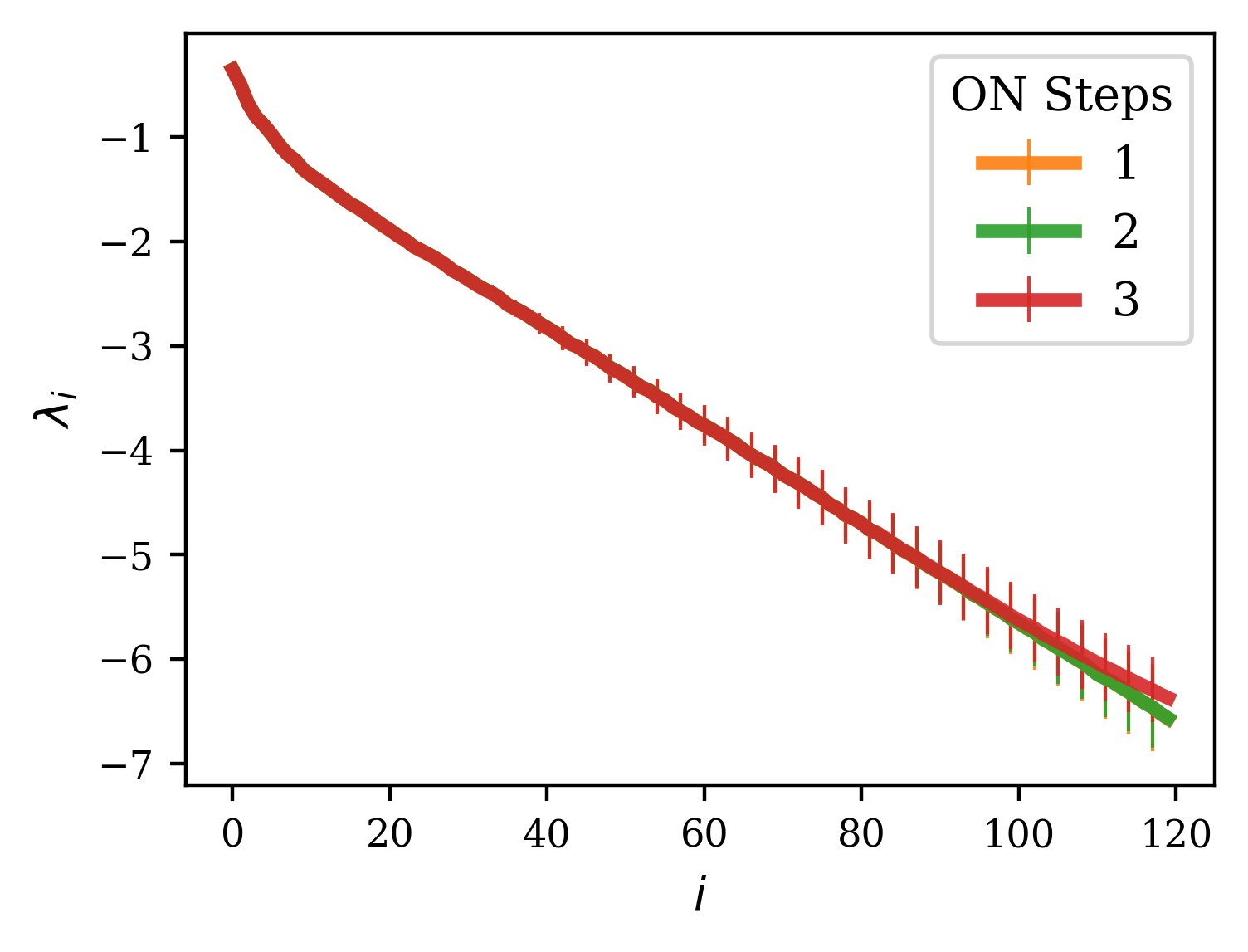}
	\end{subfigure}
\caption{{\it The effect of selection of $T_{ON}$ on the resulting spectrum}. Shown are (a) the full Lyapunov spectrum  and (b) the first half of the spectrum for a trained network with different choices of $T_{ON}$ (part of full spectrum plot indicated by gray box). Whereas the selection of $T_{ON}$ clearly impacts the second half of the spectrum, the first half remains nearly identical for all these selections of $T_{ON}$.
}
\label{fig:T_ON}
\vspace*{-3mm}
\end{figure}

A further improvement can be made by not taking the QR decomposition every time step during the calculation of the exponents, but instead every $T_{ON}$ steps. Since the QR step is the most expensive computation in the algorithm, the increase in speed over orthonormalizing every step is approximately $T_{ON}$. However, since increasing this interval leads to greater expansion and contraction of the vectors of \textbf{Q} before orthonormalization steps, this can lead to a spurious plateau at higher indices due to the accumulation of rounding errors ({\it c.f.} \cite{Engelken}. However, if one cares only about the first few exponents, this effect is negligible for reasonable selections of $T_{ON}$. In Fig. \ref{fig:T_ON}, we show how increasing $T_{ON}$ flattens out the spectrum beyond a certain index, but that the beginning of the spectrum remains unaffected by the selection of $T_{ON}$.

\setcounter{algocf}{1}
\begin{algorithm}[H]
    \SetAlgoLined
    \For{$x^j$ in Ensemble}{
    initialize \textbf{h}, \textbf{Q}\\
    Warm-up \textbf{h}, \textbf{Q}\\
    \For{t = 1 $\rightarrow$ T}{
    \textbf{h} $\leftarrow$ f($\textbf{h}$, $x^j_t$)\\
    J $\leftarrow \frac{d\textbf{f}}{d\textbf{h}}$\\
    Q $\leftarrow$ J$\cdot$Q\\
    \If{t \% $T_{ON} = 0$}{
    Q, R $\leftarrow$ $qr$(Q)\\
    $\gamma_i$ += log(R$_{ii}$)
    }
    }
    $\lambda_i = \gamma_i/T$
    }
    \caption{Modified Lyapunov Exponents Calculation}
    $\lambda_i$ = mean$_j(\lambda_i^j)$
    \label{algo: lyap_mod}
\end{algorithm}
\clearpage
\section{Spectrum metrics' correlation with loss}

\begin{figure}[h!]
	\centering
    \vspace*{-1mm}
	\begin{subfigure}{.4\linewidth}
	\caption{CharRNN}
	\includegraphics[width= \linewidth]{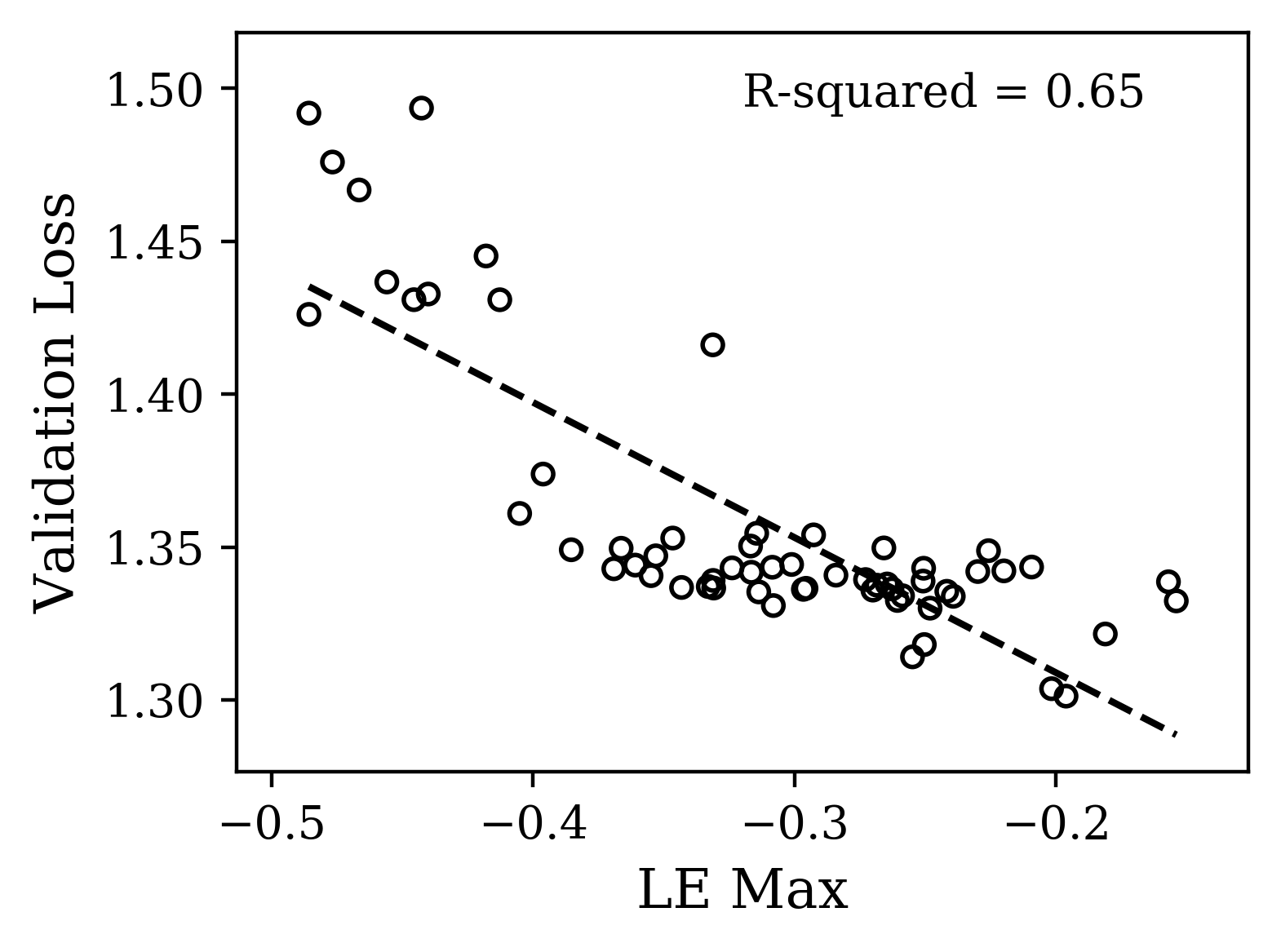}
    \end{subfigure}
    \begin{subfigure}{.4\linewidth}
    \caption{CMU Mocap}
	\includegraphics[width= \linewidth]{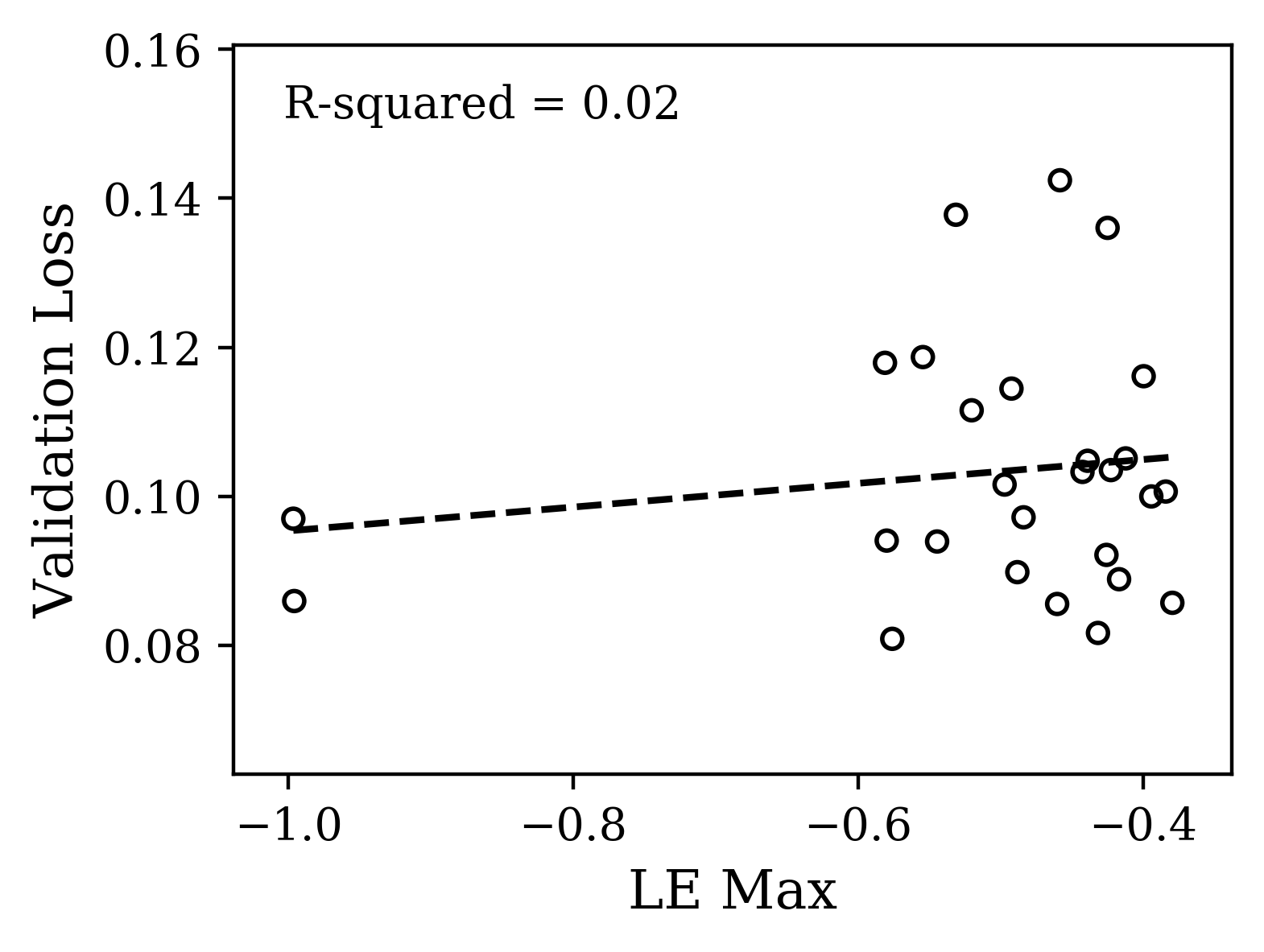}
	\end{subfigure}\\
	
	\begin{subfigure}{.4\linewidth}
    \includegraphics[width= \linewidth]{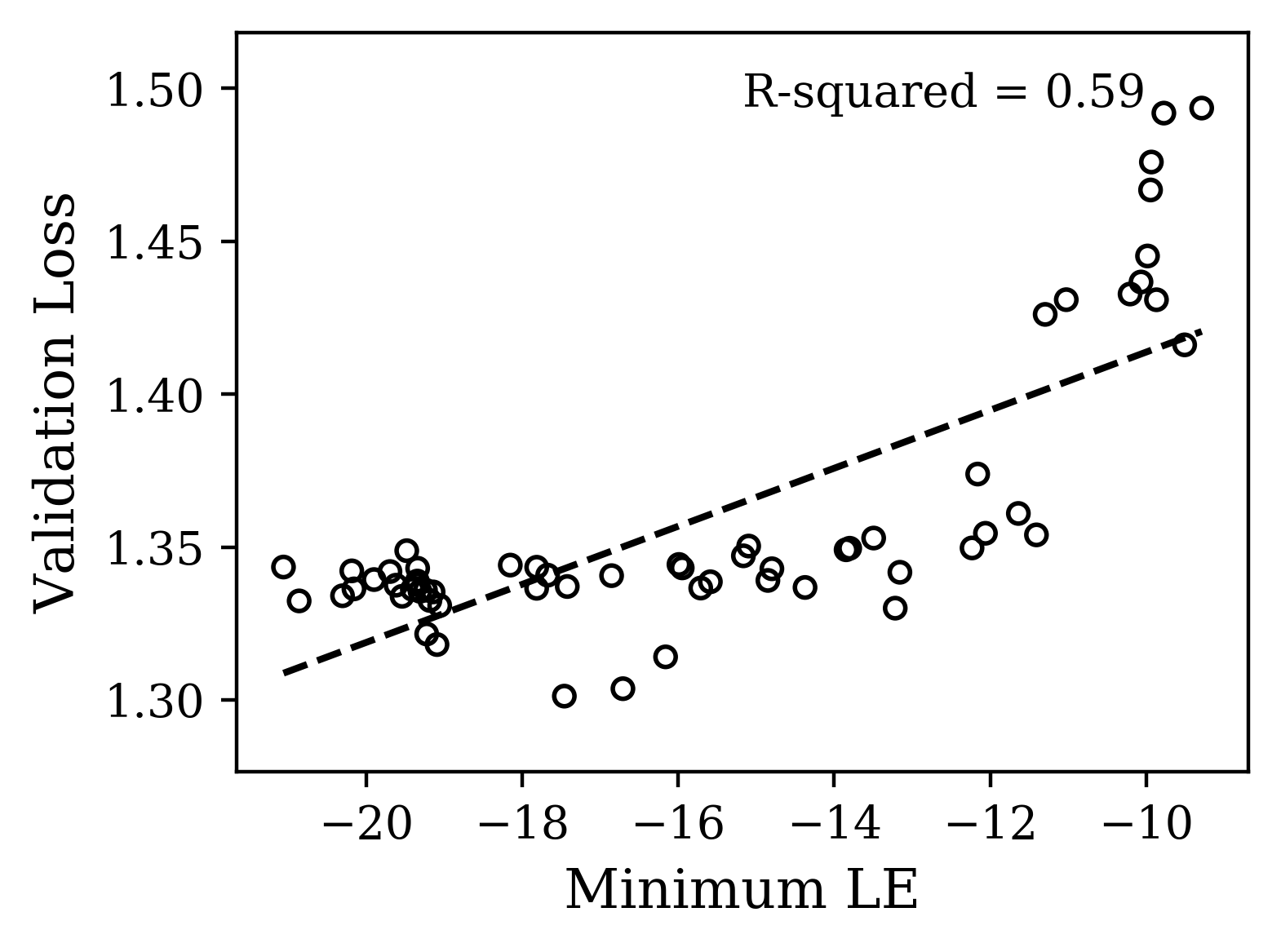}
    \end{subfigure}
    \begin{subfigure}{.4\linewidth}
	\includegraphics[width= \linewidth]{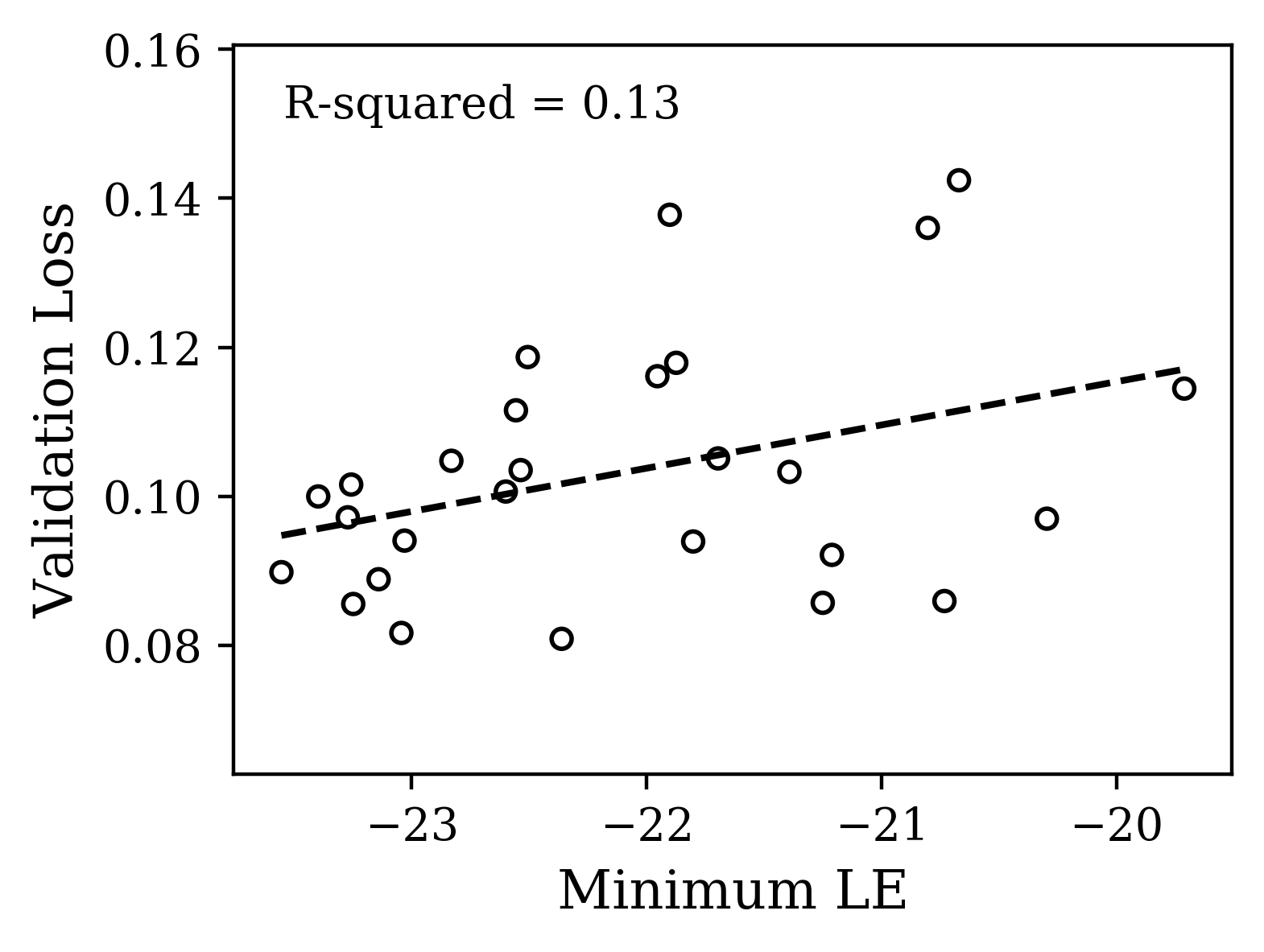}
	\end{subfigure}\\
	
	\begin{subfigure}{.4\linewidth}
    \includegraphics[width= \linewidth]{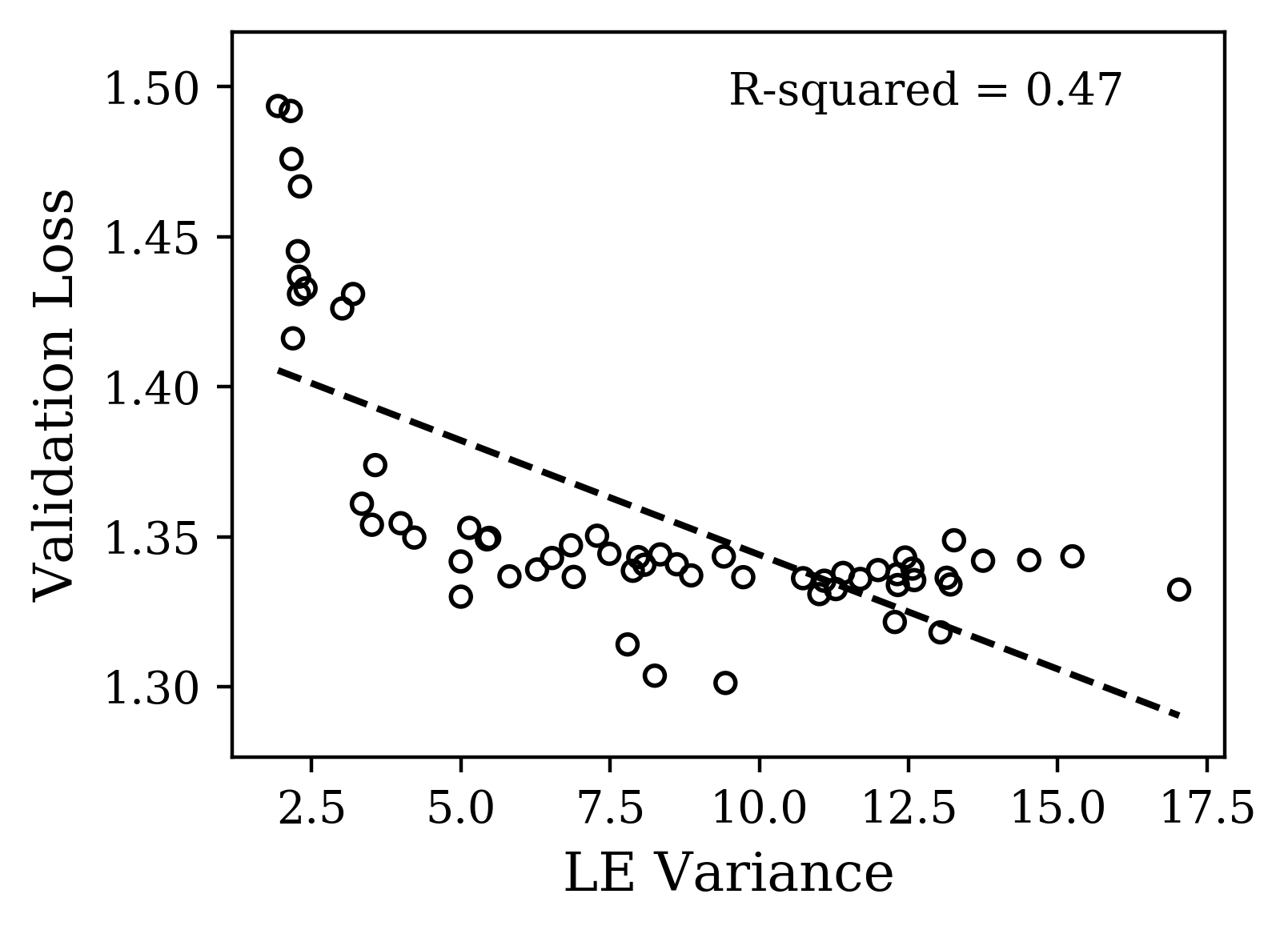}
    \end{subfigure}
    \begin{subfigure}{.4\linewidth}
	\includegraphics[width= \linewidth]{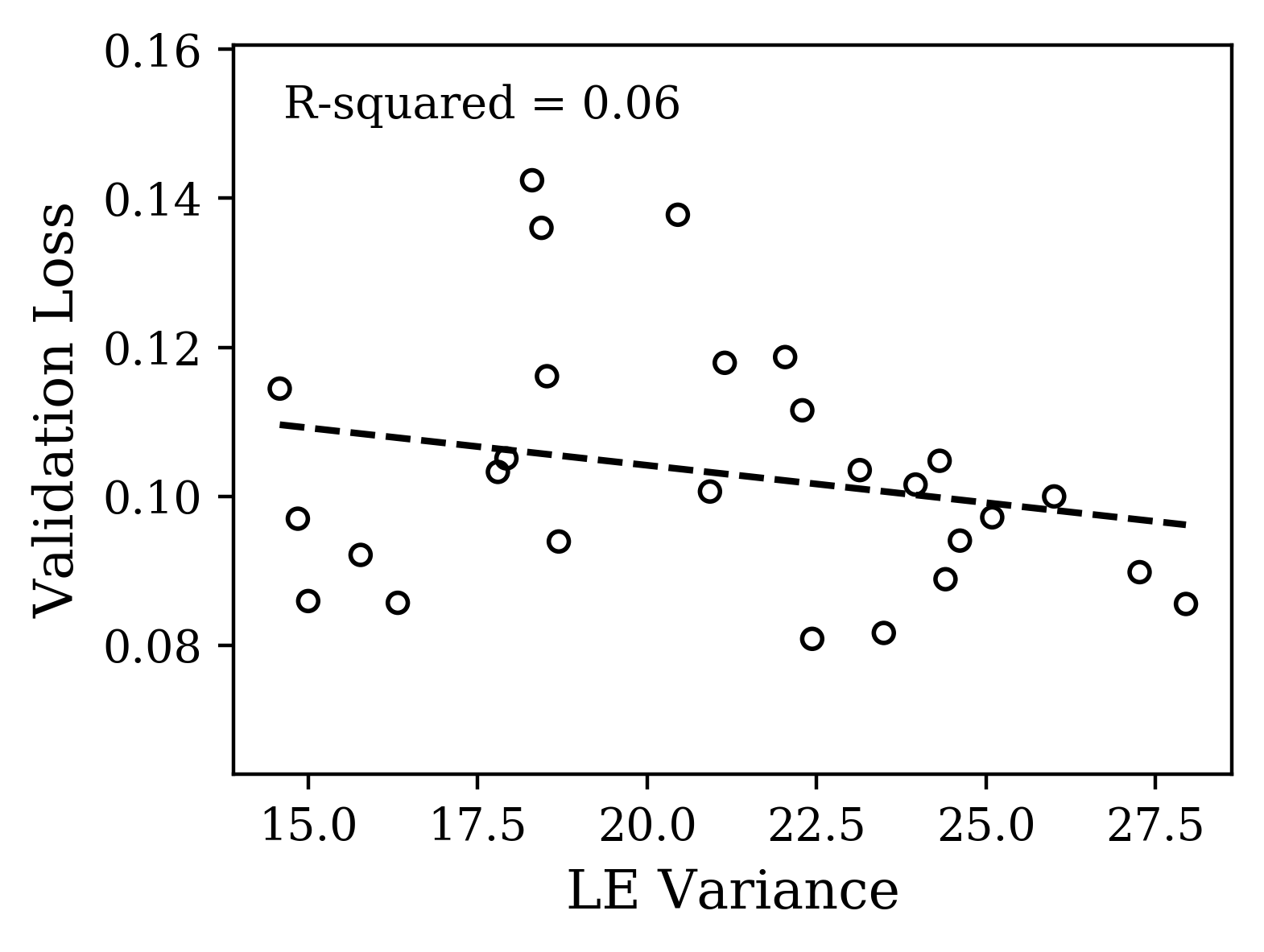}
	\end{subfigure}\\
	
	\begin{subfigure}{.4\linewidth}
    \includegraphics[width= \linewidth]{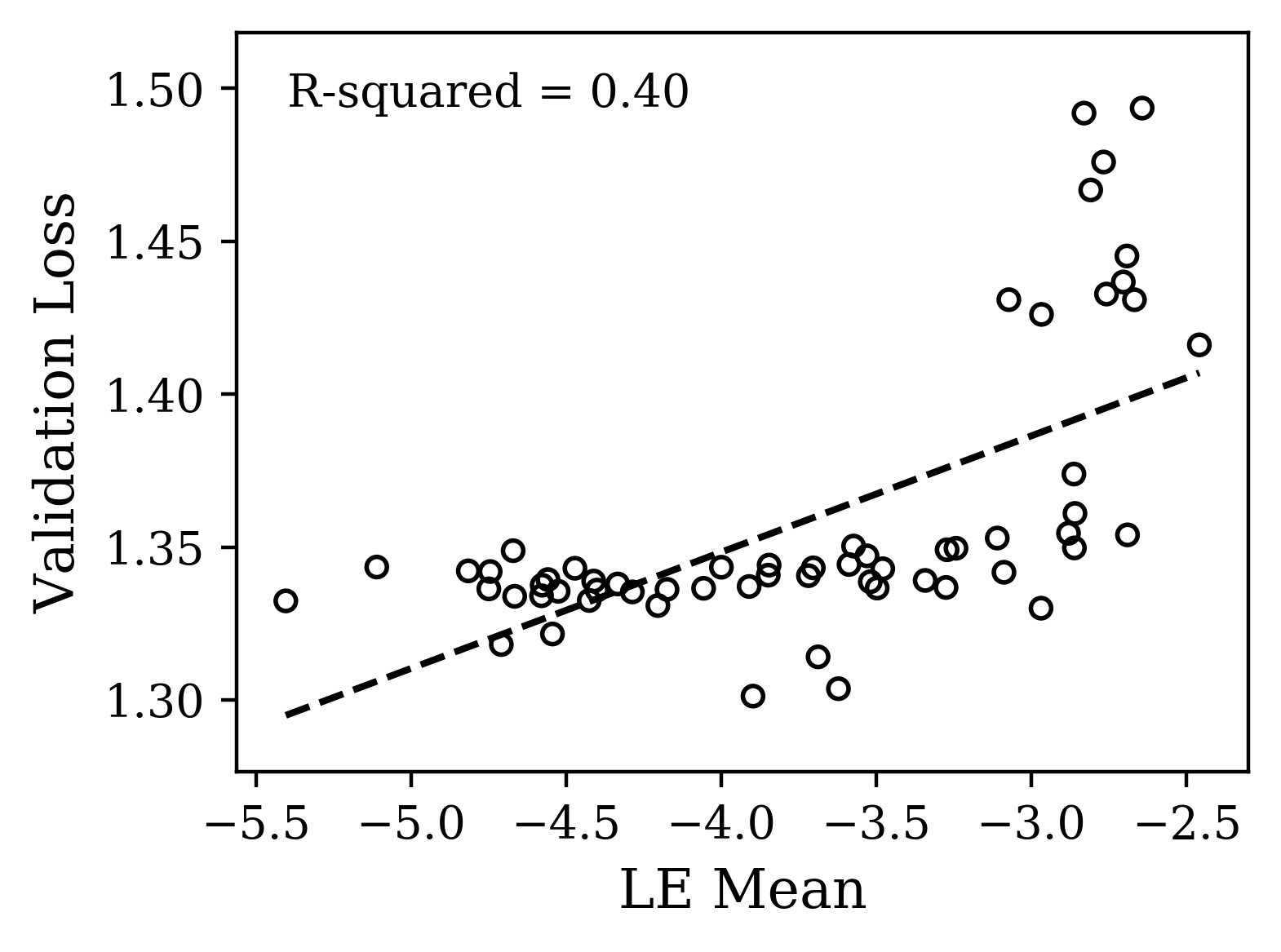}
    \end{subfigure}
    \begin{subfigure}{.4\linewidth}
	\includegraphics[width= \linewidth]{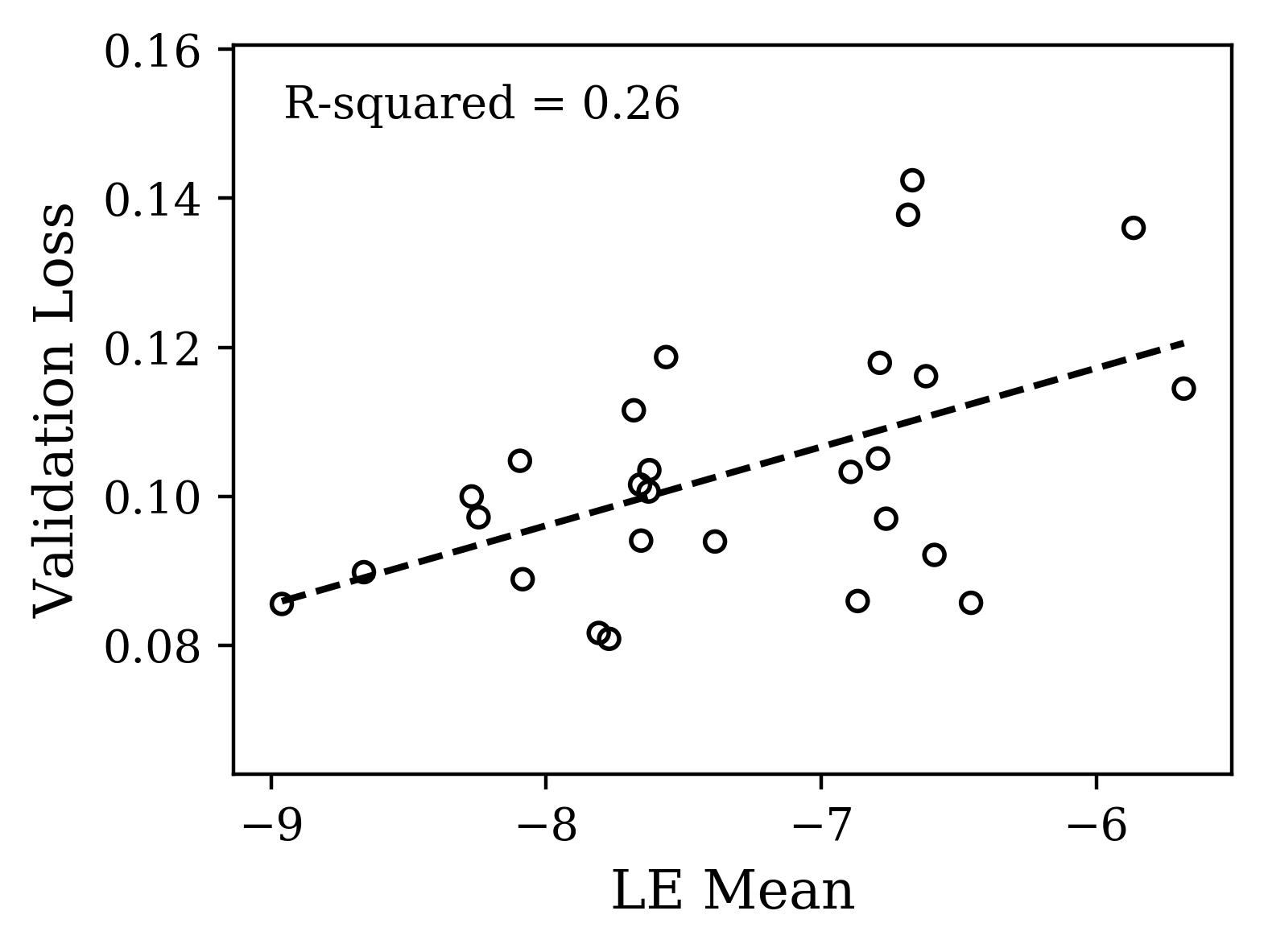}
	\end{subfigure}\\

\caption{{\it Lyapunov spectrum statistics against validation loss for both tasks}. The results for the CharRNN task are shown on the left and CMU Mocap are shown on the right. Each row represents a different statistic of the spectrum.}
\end{figure}

\end{document}